Florentin Smarandache

# Neutrosophic Overset, Neutrosophic Underset, and Neutrosophic Offset.

Similarly for Neutrosophic Over-/Under-/Off-Logic, Probability, and Statistics

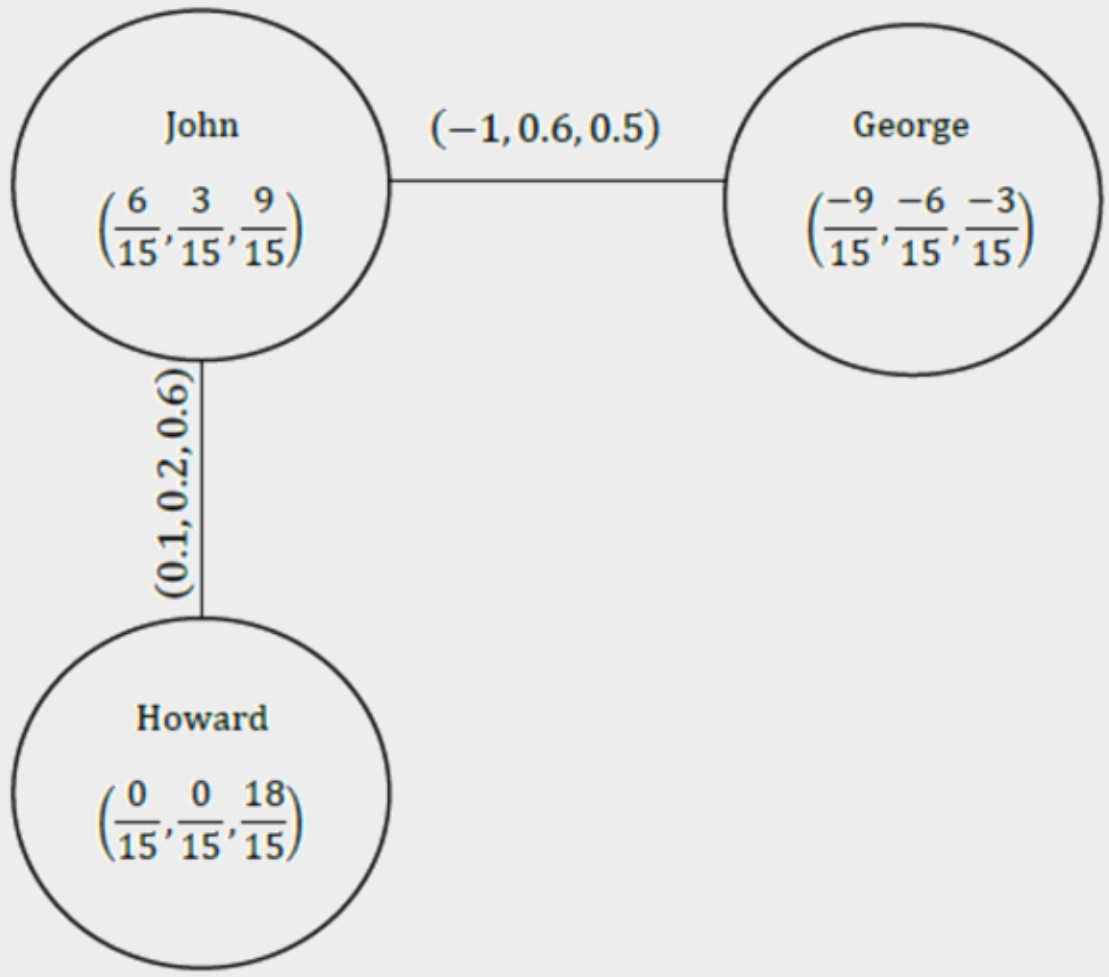

Florentin Smarandache

# Neutrosophic Overset, Neutrosophic Underset, and Neutrosophic Offset

## Similarly for Neutrosophic Over-/Under-/Off- Logic, Probability, and Statistics

***Peer-Reviewers:***

Florentin Smarandache

# Neutrosophic Overset, Neutrosophic Underset, and Neutrosophic Offset

## Similarly for Neutrosophic Over-/Under-/Off- Logic, Probability, and Statistics

# Contents

# Degree of Membership Greater Than 1 and Degree of Membership Less Than 0

(Preface)

*Neutrosophic Over--/Under-/Off-Set and -Logic* [1] were defined for the first time by the author in 1995 and presented to various international and national conferences and seminars [14-35] between 1995-2016 and first time published [1-9, 13] in 2007. They are totally different from other sets/logics/probabilities/statistics.

We extended the neutrosophic set respectively to *Neutrosophic Overset* {when some neutrosophic component is > 1}, *Neutrosophic Underset* {when some neutrosophic component is < 0}, and to *Neutrosophic Offset* {when some neutrosophic components are off the interval [0, 1], i.e. some neutrosophic component > 1 and other neutrosophic component < 0}.

This is no surprise with respect to the classical fuzzy set/ logic, intuitionistic fuzzy set/ logic, or classical/ imprecise probability, where the values are not allowed outside the interval [0, 1], since our real-world has numerous examples and applications of over-/under-/off-neutrosophic components.

*Example:*

In a given company a full-time employer works 40 hours per week. Let's consider the last week period.

Helen worked part-time, only 30 hours, and the other 10 hours she was absent without payment; hence, her membership degree was 30/40 = 0.75 < 1.

John worked full-time, 40 hours, so he had the membership degree 40/40 = 1, with respect to this company.

But George worked overtime 5 hours, so his membership degree was (40+5)/40 = 45/40 = 1.125 > 1. Thus, we need to make distinction between employees who work overtime, and those who work full-time or part-time. That's why we need to associate a degree of membership strictly greater than 1 to the overtime workers.

Now, another employee, Jane, was absent without pay for the whole week, so her degree of membership was 0/40 = 0.

Yet, Richard, who was also hired as a full-time, not only didn't come to work last week at all (0 worked hours), but he produced, by accidentally starting a devastating fire, much damage to the company, which was estimated at a value half of his salary (i.e. as he would have gotten for working 20 hours that week). Therefore, his membership degree has to be less that Jane's (since Jane produced no damage). Whence, Richard's degree of membership, with respect to this company, was - 20/40 = - 0.50 < 0. Thus, we need to make distinction between employees who produce damage, and those who produce profit, or produce neither damage no profit to the company.

Therefore, the membership degrees > 1 and < 0 are real in our world, so we have to take them into consideration.

Then, similarly, the Neutrosophic Logic / Measure / Probability / Statistics etc. were extended to respectively *Neutrosophic Over- / Under- / Off -Logic, -Measure, -Probability, -Statistics* etc. [Smarandache, 2007].

Many practical obvious examples are presented in this book, in order to show that in our everyday life we continuously deal with neutrosophic over-/under-/off-theory and applications.

Feedback from the readers about this very new *neutrosophic over- / under- / off- (–set, –logic, –measure, –probability, –statistics)* and their applications is very welcomed by the author.

Prof. Florentin Smarandache, Ph D
University of New Mexico
Mathematics & Science Department
705 Gurley Ave., Gallup, NM 87301, USA
E-mail: *smarand@unm.edu*
*http://fs.gallup.unm.edu/FlorentinSmarandache.htm*

# New Neutrosophic Terminology

We introduce several new scientific notions in the domain of Neutrosophic Theory and Its Applications, coined now for the first time upon the best of our knowledge, created by juxtaposition of two words, i.e.

a) inserting the *prefix* "over", "under", or "off"
b) in front of a *noun*, such as:
   - "membership", "indeterminate-membership", "nonmembership";
   - or "truth", "indeterminacy", "falsehood";
   - or "element";
   - or "graph", "matrix" etc.
   - or "set", "logic", "measure", "topology", "probability", "statistics" etc.

## Etymology

- *Overtruth* is like over-confidence [believing too much in something], over-estimation, overwhelming [much above the limit], overcharging, overdose, overdeveloped, overproduction, overdone, overbidding, overheating, overexciting etc.

So, overtruth (overtrue) means: over the truth, above the truth, more than the truth (i.e. percentage of truth > 100%).

- *Overmembership* means similarly: more than full-time membership, i.e. over-time membership (degree of membership > 100%).

- *Undertruth* is like under-confidence, under-estimation, undercharging, under-dose, underdeveloped,

underproduction, underdone, underbidding, under-heating, under-exciting etc.

So, undertruth (undertrue) means: under the truth, below the truth (i.e. percentage of truth < 0%).

- *Undermembership* means similarly: under the membership degree, i.e. negative membership (degree of membership < 0%).

- *Offtruth* is like off-confidence, off-estimation, off-production, off-side, off-stage, off-key, off-load, etc.

So, offtruth (offtrue) means: over the truth or under the truth, above the truth and below the truth (i.e. a percentage of truth > 100% and one < 0%).

- *Offmembership* means similarly: over-time membership degree, or below membership degree (i.e. a degree of membership > 100%, and a degree of membership < 0%).

Similarly for the: *overindeterminacy, overfalsehood (overfalsity); underindeterminacy, underfalsehood (under-falsity); offindeterminacy, offfalsehood (offfalsity).*

OVER.

We define the:

*neutrosophic overelement*, which is an element that has at least one of its neutrosophic components T, I, F that is > 1.

Whence, we define the:

*neutrosophic overgraph,*

*neutrosophic overmatrix,*

and especially the

*neutrosophic overset, neutrosophic overmeasure, neutrosophic overtopology, neutrosophic overprobability, neutrosophic overstatistics,*

which are mathematical objects or structures that contain at least one neutrosophic overelement.

UNDER.

We define the:

*neutrosophic underelement*, which is an element that has at least one of its neutrosophic components T, I, F that is < 0.

Whence, we define the:

*neutrosophic undergraph,*
*neutrosophic undermatrix,*

and especially the

*neutrosophic underset, neutrosophic undermeasure, neutrosophic undertopology, neutrosophic underprobability, neutrosophic understatistics*

which are mathematical objects or structures that contain as least one neutrosophic underelement.

OFF.

We define the:

*neutrosophic offelement*, which is an element that has at least two of its neutrosophic components T, I, F such that one is > 0 and one is < 0.

Whence, we define the:

*neutrosophic offgraph,*
*neutrosophic offmatrix,*

and especially the

*neutrosophic offset, neutrosophic offmeasure, neutrosophic offtopology, neutrosophic offprobability, neutrosophic offstatistics*

which are mathematical objects or structures that contain as least one neutrosophic offelement, or at least one neutrosophic overelement and one neutrosophic underelement.

# Introduction

The idea of membership degree >1, for an element with respect to a set, came to my mind when I started teaching and doing scientific presentations at several Colleges and Universities in The United States since 1995.

A student was considered full-time for a semester if he or she enrolled in five classes. Therefore, his/her membership was 1 or T(student) = 1.

But there were students enrolled in six classes as well (overloaded). Then a twinkle sparked in my mind: I thought it was normal to consider such student's membership degree greater than 1, or T(overload student) = $\frac{6}{5} = 1.2 > 1$.

Surely, this was in contradiction with the orthodoxy that the crisp membership degree of an element with respect to a set has to be $\leq 1$.

I dug more into the problem and looked for other examples and applications from our everyday life. I did not want to stick with the abstractness of the mathematics, but to be inspired from our concrete reality.

I was even more shocked when I discovered examples of membership degree < 0 of an element with respect to a set.

For example, let's consider the set of spy agents of a country against an enemy country.

A full-time spy, working only for his country, has the degree of membership equals to 1 with respect to the set of spy agents in his country. He is productive.

But a double-agent, that leaks highly classified information to the enemy country, while to his country he provides false information about the enemy country, produces much damage to his country (he is counter-productive), hence he has a negative membership degree with respect to the set of spy agents of his country, since he actually belongs to the set of spy agents of the enemy country, thus $T$(double-agent) < 0. He is counter-productive.

At that time, I was also struggling to convince people about the viability of neutrosophic set and neutrosophic logic, i.e. that the sum of the crisp neutrosophic components

$T + I + F$ can exceed 1,

even more, that the sum

$T + I + F$ can be extended to 3

when the three sources that provide us information about $T$ (degree of membership / truth), $I$ (degree of indeterminacy regarding the membership / truth), and respectively F (degree of nonmembership / falsehood) are independent, while fuzzy set and fuzzy logic, intuitionistic fuzzy set and intuitionistic fuzzy logic, in addition of classical probability did not allow this.

It took me about three years (1995-1998) to little by little move the thinking out from the routine of upper bound = 1.

I was criticized that I "ignored the elementary things about probability," i.e. that the sum of space probabilities is equal to 1. But this is true for objective classical probability, not for subjective probability.

"Neutrosophic" means based on three components $T$, $I$, and $F$; and "offset" means behind/beside the set on both sides of the interval [0, 1], over and under, more and less, supra and below, out of, off the set. Similarly, for "offlogic", "offmeasure", "offprobability", "offstatistics" etc.

It is like a pot with boiling liquid, on a gas stove, when the liquid swells up and leaks out of pot. The pot (the interval [0, 1]) can no longer contain all liquid (i.e., all neutrosophic truth / indeterminate / falsehood values), and therefore some of them fall out of the pot (i.e., one gets neutrosophic truth / indeterminate / falsehood values which are > 1), or the pot cracks on the bottom and the liquid pours down (i.e., one gets neutrosophic truth / indeterminate / falsehood values which are < 0).

Mathematically, they mean getting values off the interval [0, 1].

The American aphorism "think outside the box" has a perfect resonance to the neutrosophic offset, where the box is the interval [0, 1], yet values outside of this interval are permitted.

# 1. Definition of Single-Valued Neutrosophic Overset

Let $\mathcal{U}$ be a universe of discourse and the neutrosophic set $A_1 \subset \mathcal{U}$.

Let T(x), I(x), F(x) be the functions that describe the degrees of membership, indeterminate-membership, and nonmembership respectively, of a generic element $x \in \mathcal{U}$, with respect to the neutrosophic set $A_1$:

$$T(x), I(x), F(x) : \mathcal{U} \rightarrow [0, \Omega] \quad (1)$$

where $0 < 1 < \Omega$, and $\Omega$ is called overlimit,

$$T(x), I(x), F(x) \in [0, \Omega]. \quad (2)$$

A **Single-Valued Neutrosophic Overset** $A_1$ is defined as:

$$A_1 = \{(x, <T(x), I(x), F(x)>), x \in \mathcal{U}\}, \quad (3)$$

such that there exists at least one element in $A_1$ that has at least one neutrosophic component that is > 1, and no element has neutrosophic components that are < 0.

For **example**: $A_1 = \{(x_1, <1.3, 0.5, 0.1>), (x_2, <0.2, 1.1, 0.2>)\}$, since $T(x_1) = 1.3 > 1$, $I(x_2) = 1.1 > 1$, and no neutrosophic component is < 0.

Also $O_2 = \{(a, <0.3, -0.1, 1.1>)\}$, since $I(a) = - 0.1 < 0$ and $F(a) = 1.1 > 1$.

## 2. Definition of Single Valued Neutrosophic Underset

Let $\mathcal{U}$ be a universe of discourse and the neutrosophic set $A_2 \subset \mathcal{U}$.

Let T(x), I(x), F(x) be the functions that describe the degrees of membership, indeterminate-membership, and nonmembership respectively, of a generic element $x \in \mathcal{U}$, with respect to the neutrosophic set $A_2$:

$$T(x), I(x), F(x) : \mathcal{U} \to [\Psi, 1] \quad (4)$$

where $\Psi < 0 < 1$, and $\Psi$ is called underlimit,

$$T(x), I(x), F(x) \in [\Psi, 1]. \quad (5)$$

A **Single-Valued Neutrosophic Underset** $A_2$ is defined as:

$$A_2 = \{(x, <T(x), I(x), F(x)>), x \in \mathcal{U}\}, \quad (6)$$

such that there exists at least one element in $A_2$ that has at least one neutrosophic component that is < 0, and no element has neutrosophic components that are > 1.

For **example**: $A_2 = \{(x_1, <-0.4, 0.5, 0.3>), (x_2, <0.2, 0.5, -0.2>)\}$, since $T(x_1) = -0.4 < 0$, $F(x_2) = -0.2 < 0$, and no neutrosophic component is > 1.

# 3. Definition of Single-Valued Neutrosophic Offset

Let $\mathcal{U}$ be a universe of discourse and the neutrosophic set $A_3 \subset \mathcal{U}$.

Let T(x), I(x), F(x) be the functions that describe the degrees of membership, indeterminate-membership, and nonmembership respectively, of a generic element $x \in \mathcal{U}$, with respect to the set $A_3$:

$$T(x), I(x), F(x) : \mathcal{U} \rightarrow [\Psi, \Omega] \qquad (7)$$

where $\Psi < 0 < 1 < \Omega$, and $\Psi$ is called underlimit, while $\Omega$ is called overlimit,

$$T(x), I(x), F(x) \in [\Psi, \Omega]. \qquad (8)$$

A **Single-Valued Neutrosophic Offset** $A_3$ is defined as:

$$A_3 = \{(x, <T(x), I(x), F(x)>), x \in \mathcal{U}\}, \qquad (9)$$

such that there exist some elements in $A_3$ that have at least one neutrosophic component that is > 1, and at least another neutrosophic component that is < 0.

For **examples**: $A_3 = \{(x_1, <1.2, 0.4, 0.1>), (x_2, <0.2, 0.3, -0.7>)\}$, since $T(x_1) = 1.2 > 1$ and $F(x_2) = -0.7 < 0$.

Also $B_3 = \{(a, <0.3, -0.1, 1.1>)\}$, since $I(a) = -0.1 < 0$ and $F(a) = 1.1 > 1$.

## Example of Overindeterminacy

At the University Alpha, the norm of a full-time student is 15 credit hours, but the students are allowed to enroll in overload up to 18 credit hours.

The student Edward has enrolled in 18 credit hours, but his enrollment is pending because of financial aid.

Therefore, Edward's membership to the University Alpha is Edward $\left(0, \frac{18}{15}, 0\right)$ = Edward $(0, 1.2, 0)$ , i.e. overindeterminate $(1.2 > 1)$.

## Example of Relative Membership

Universities Alpha and Beta fight for attracting students. If University Alpha succeeds to attract the student Marcel to enroll in, let's say, 6 credit hours, then Marcel's membership with respect to the University Alpha is + $\frac{6}{15} = +\,0.4$ (positive), while Marcel's membership with respect to the University Beta is $-\frac{6}{15} = -0.4$ (negative), since it was lost to Alpha, which is Beta's competitor/rival.

Suppose there exists a third university, University Gamma, in the same city which does not compete against University Alpha or University Beta because it has a different profile of offered courses. Then Marcel's membership with respect to Gamma is $\frac{0}{15} = 0$ (zero), since he enrolling in Alpha or Beta does not affect University Gamma's enrollment.

# Example of Underindeterminacy

Similarly, if Marcel, in addition to the 6 credit hours enrolled at the University Alpha, has enrolled at the concurrent University Beta in 3 credit hours, which are pending.

Hence Marcel's membership with respect to Alpha is:

$$\left(\frac{6}{15}, \frac{-3}{15}, \frac{18-6}{15}\right)_{Alpha} = (0.4, -0.2, 0.8)_{Alpha}.$$

Marcel's positive indeterminacy with respect to Beta, $+\frac{3}{15}$, is translated to negative indeterminacy with respect to Alpha, $-\frac{3}{15}$, since if the pending is resolved the indeterminacy $\frac{3}{15}$ with respect to Beta becomes membership $\frac{3}{15}$ with respect to Beta, which means $-\frac{3}{15}$ membership with respect to Alpha.

# Example of Overnonmembership

At the University Beta, where the full-time norm for a student in 15 credit hours, and the overload is allowed up to 21 credit hours, a student, Frederic, has enrolled in 3 credit units.

His membership with respect to Beta is:

Frederic $\left(\frac{3}{15}, \frac{0}{15}, \frac{21-3}{15}\right)_{Beta}$ = Frederic $(0.2, 0, 1.2)_{Beta}$, since he had the possibility in enroll in $21 - 3 = 18$ more credit hours.

His nonmebership with respect to Beta is 1.2 > 1.

## Example of Undernonmembership

At the University Alpha the full-time norm for a student is 15 credit hours, and the maximum overload allowed is up to 18 credit hours. Therefore:

$-\frac{18}{15} \leq T, I, F \leq \frac{18}{15}$, or

$-1.2 \leq \ T, I, F \leq 1.2.$

Helen, a brilliant student, is enrolled in 18 credit hours. Therefore, one has Helen$\left(\frac{18}{15}, \frac{0}{15}, \frac{0}{15}\right)_{Alpha} = (1.2, 0, 0)_{Alpha}$.

But, due to her high performance in studying, the Alpha's President and Provost approve her, exceptionally, to enroll, additionally, in an honorary course of 2 credit hours. Since her membership could become $\frac{18+2}{5} = \frac{20}{15} \simeq 1.33$, which is off the interval [-1.2, 1.2], instead of considering her overmembership to the University Alpha $\left(\frac{20}{15}, \frac{0}{15}, \frac{0}{15}\right)$ one moves her positive 2-credit hours membership as negative 2-credit hours nonmembership, therefore one gets $\left(\frac{18}{15}, \frac{0}{15}, \frac{-2}{15}\right) = (1.2, 0, -0.13)$ , and now all the three neutrosophic components are within the frame [-1.2, +1.2].

Surely, as a precedent, the University Alpha's Board of Regents may discuss for the future to extend the maximum overload up to 20 credit hours. And, as a consequence, in this new frame, Helen's membership would be allowed to be $\left(\frac{20}{15}, \frac{0}{15}, \frac{0}{15}\right)$.

# Future Research

As possible future research, for interested reader, it will be the case when the classical unit interval [0, 1] is not included in at least one of the off-set intervals

$[\Psi_T, \Omega_T], [\Psi_I, \Omega_I], [\Psi_F, \Omega_F]$.

For example, the case when a lower threshold is > 0 or a upper threshold is < 1.

# A Simple Example on Upper and Lower Thresholds

The "Andromeda" ship does cruises from Ushuaia (South Argentina) to Antarctica for the price of 15k per tourist. Therefore, a person paying 15k is considered a full-time tourist. But, due to the world crisis, the "Andromeda" ship crew gets no costumer!

Then, the ship's captain decides to make a discount of 20% in order for not losing everything. Therefore, the tourist's membership (appurtenance to the cruise from a financial point of view) was at the beginning [0, 1] corresponding to [0, 15k]. But later it became:

$$\left[0, \frac{15k-(20\%\ of 15k))}{15k}\right] = \left[0, \frac{15k-3k}{15k}\right] = [0, 0.8].$$

Hence, the upper threshold of membership is not classical (1), but less (0.8).

Although a discount has been made, still not enough passengers on the ship. The, the ship captain, in order to fill in all remaining places on ship, allows for the last passengers up to 50% discount.

So, the lower threshold is not zero (0), but $\frac{50\%\ of\ 15k}{15k} = 0.5$. Whence, the interval of membership of the tourists / passengers becomes [0.5, 0.8], not [0, 1].

# 4. Single-Valued Neutrosophic Overset / Underset / Offset Operators

Let $\mathcal{U}$ be a universe of discourse and A = {(x, <$T_A$(x), $I_A$(x), $F_A$(x)>), x ∈ $\mathcal{U}$} and B = {(x, <$T_B$(x), $I_B$(x), $F_B$(x)>), x ∈ $\mathcal{U}$} be two single-valued neutrosophic oversets / undersets / offsets.

$T_A(x), I_A(x), F_A(x), T_B(x), I_B(x), F_B(x): \mathcal{U} \to [\Psi,\Omega]$ (10)

where $\Psi \leq 0 < 1 \leq \Omega$, and $\Psi$ is called underlimit, while $\Omega$ is called overlimit,

$T_A(x), I_A(x), F_A(x), T_B(x), I_B(x), F_B(x) \in [\Psi,\Omega]$ . (11)

We take the inequality sign ≤ instead of < on both extremes above, in order to comprise all three cases: overset {when $\Psi = 0$, and $1 < \Omega$}, underset {when $\Psi < 0$, and $1 = \Omega$}, and offset {when $\Psi < 0$, and $1 < \Omega$}.

## Single-Valued Neutrosophic Overset / Underset / Offset Union

Then A ∪ B = {(x, <max{$T_A$(x), $T_B$(x)}, min{$I_A$(x), $I_B$(x)}, min{$F_A$(x), $F_B$(x)}>), x∈ U}. (12)

## Single-Valued Neutrosophic Overset / Underset / Offset Intersection

Then A ∩ B = {(x, <min{$T_A$(x), $T_B$(x)}, max{$I_A$(x), $I_B$(x)}, max{$F_A$(x), $F_B$(x)}>), x∈ U}. (13)

## Single-Valued Neutrosophic Overset / Underset / Offset Complement

The neutrosophic complement of the neutrosophic set A is C(A) = {(x, <$F_A$(x), $\Psi + \Omega - I_A(x)$, $T_A$(x)>), x ∈ U}. (14)

# 5. Definition of Interval-Valued Neutrosophic Overset

Let $\mathcal{U}$ be a universe of discourse and the neutrosophic set $A_1 \subset \mathcal{U}$.

Let T(x), I(x), F(x) be the functions that describe the degrees of membership, indeterminate-membership, and nonmembership respectively, of a generic element $x \in \mathcal{U}$, with respect to the neutrosophic set $A_1$:

$$T(x), I(x), F(x) : \mathcal{U} \to P([0, \Omega]), \quad (15)$$

where $0 < 1 < \Omega$, and $\Omega$ is called overlimit,

$$T(x), I(x), F(x) \subseteq [0, \Omega], \quad (16)$$

and $P([0, \Omega])$ is the set of all subsets of $[0, \Omega]$.

An Interval-Valued Neutrosophic Overset $A_1$ is defined as:

$$A_1 = \{(x, <T(x), I(x), F(x)>), x \in \mathcal{U}\}, \quad (17)$$

such that there exists at least one element in $A_1$ that has at least one neutrosophic component that is partially or totally above 1, and no element has neutrosophic components that is partially or totally below 0.

For **example**: $A_1 = \{(x_1, <(1, 1.4], 0.1, 0.2>), (x_2, <0.2, [0.9, 1.1], 0.2>)\}$, since $T(x_1) = (1, 1.4]$ is totally above 1, $I(x_2) = [0.9, 1.1]$ is partially above 1, and no neutrosophic component is partially or totally below 0.

# 6. Definition of Interval-Valued Neutrosophic Underset

Let $\mathcal{U}$ be a universe of discourse and the neutrosophic set $A_2 \subset \mathcal{U}$.

Let T(x), I(x), F(x) be the functions that describe the degrees of membership, indeterminate-membership, and nonmembership respectively, of a generic element x ∈ U, with respect to the neutrosophic set $A_2$:

$T(x), I(x), F(x) : \mathcal{U} \rightarrow [\Psi, 1]$, (18)

where $\Psi < 0 < 1$, and $\Psi$ is called underlimit,

$T(x), I(x), F(x) \subseteq [\Psi, 1]$ , (19)

and $P([\Psi, 1])$ is the set of all subsets of $[\Psi, 1]$ .

An Interval-Valued Neutrosophic Underset $A_2$ is defined as:

$A_2 = \{(x, <T(x), I(x), F(x)>), x \in \mathcal{U} \}$, (20)

such that there exists at least one element in $A_2$ that has at least one neutrosophic component that is partially or totally below 0, and no element has neutrosophic components that are partially or totally above 1.

For **example**: $A_2 = \{(x_1, <(-0.5,-0.4), 0.6, 0.3>), (x_2, <0.2, 0.5, [-0.2, 0.2]>)\}$, since $T(x_1) = (-0.5, -0.4)$ is totally below 0, $F(x_2) = [-0.2, 0.2]$ is partially below 0, and no neutrosophic component is partially or totally above 1.

# 7. Definition of Interval-Valued Neutrosophic Offset

Let $\mathcal{U}$ be a universe of discourse and the neutrosophic set $A_3 \subset \mathcal{U}$.

Let T(x), I(x), F(x) be the functions that describe the degrees of membership, indeterminate-membership, and nonmembership respectively, of a generic element $x \in U$, with respect to the set $A_3$:

$$T(x), I(x), F(x) : \mathcal{U} \rightarrow P([\Psi, \Omega]), \qquad (21)$$

where $\Psi < 0 < 1 < \Omega$, and $\Psi$ is called underlimit, while $\Omega$ is called overlimit,

$$T(x), I(x), F(x) \subseteq [\Psi, \Omega], \qquad (22)$$

and $P([\Psi, \Omega])$ is the set of all subsets of $[\Psi, \Omega]$.

An Interval-Valued Neutrosophic Offset $A_3$ is defined as:

$$A_3 = \{(x, <T(x), I(x), F(x)>), x \in \mathcal{U}\}, \qquad (23)$$

such that there exist some elements in $A_3$ that have at least one neutrosophic component that is partially or totally abive 1, and at least another neutrosophic component that is partially or totally below 0.

For **examples**: $A_3 = \{(x_1, <[1.1, 1.2], 0.4, 0.1>), (x_2, <0.2, 0.3, (-0.7, -0.3)>)\}$, since $T(x_1) = [1.1, 1.2]$ that is totally above 1, and $F(x_2) = (-0.7, -0.3)$ that is totally below 0.

Also $B_3 = \{(a, <0.3, [-0.1, 0.1], [1.05, 1.10]>)\}$, since $I(a) = [-0.1, 0.1]$ that is partially below 0, and $F(a) = [1.05, 1.10]$ that is totally above 1.

# 8. Definition of Interval-Valued Neutrosophic Overset Operators

Let $\mathcal{U}$ be a universe of discourse and A = {(x, <$T_A$(x), $I_A$(x), $F_A$(x)>), x ∈ U} and B = {(x, <$T_B$(x), $I_B$(x), $F_B$(x)>), x ∈ U} be two interval-valued neutrosophic oversets / undersets / offsets.

$$T_A(x), I_A(x), F_A(x), T_B(x), I_B(x), F_B(x): \mathcal{U} \to P([\Psi,\Omega]), \quad (24)$$

where $P([\Psi,\Omega])$ means the set of all subsets of $[\Psi,\Omega]$,

and $T_A(x), I_A(x), F_A(x), T_B(x), I_B(x), F_B(x) \subseteq [\Psi,\Omega]$,

with $\Psi \leq 0 < 1 \leq \Omega$, and $\Psi$ is called underlimit, while $\Omega$ is called overlimit.

We take the inequality sign ≤ instead of < on both extremes above, in order to comprise all three cases: overset {when $\Psi = 0$, and $1 < \Omega$}, underset {when $\Psi < 0$, and $1 = \Omega$}, and offset {when $\Psi < 0$, and $1 < \Omega$}.

## Interval-Valued Neutrosophic Overset / Underset / Off Union

Then A∪B =

{(x, <[max{inf($T_A$(x)), inf($T_B$(x))}, max{sup($T_A$(x)), sup($T_B$(x)}],

[min{inf($I_A$(x)), inf($I_B$(x))}, min{sup($I_A$(x)), sup($I_B$(x)}],

[min{inf($F_A$(x)), inf($F_B$(x))}, min{sup($F_A$(x)), sup($F_B$(x)}]>, x ∈ U}. (25)

## Interval-Valued Neutrosophic Overset / Underset / Off Intersection

Then A∩B =
{(x, <[min{inf($T_A$(x)), inf($T_B$(x))}, min{sup($T_A$(x)), sup($T_B$(x)}],
[max{inf($I_A$(x)), inf($I_B$(x))}, max{sup($I_A$(x)), sup($I_B$(x)}],
[max{inf($F_A$(x)), inf($F_B$(x))}, max{sup($F_A$(x)), sup($F_B$(x)}]>, x ∈ U}. (26)

## Interval-Valued Neutrosophic Overset / Underset / Off Complement

The complement of the neutrosophic set A is
C(A) = {(x, <$F_A$(x), [$\Psi + \Omega$ - sup{$I_A$(x)}, $\Psi + \Omega$ - inf{$I_A$(x)}], $T_A$(x)>), x ∈ U}. (27)

# 9. Definition of Subset Neutrosophic Overset

Let $\mathcal{U}$ be a universe of discourse.

Neutrosophic Overset is a set $M_{over}$ from $\mathcal{U}$ that has at least one element (called overelement)

$z(t_{M_{over}}, i_{M_{over}}, f_{M_{over}}) \in M_{over}$

whose at least one neutrosophic component $t_{M_{over}}, i_{M_{over}}, f_{M_{over}}$ is partially or totally > 1.

For *example*, the following **overelements**:

$d(1.2, 0.4, 0)$ (overtruth, or overmembership),

$e(0.9, 1.3, 0.6)$ (overindeterminacy),

$k([0.1, 0.4], (0.5, 0.7), (0.9, 1.6])$ (overfalsity, or overnonmembership).

Therefore, a neutrosophic overset has elements with neutrosophic components strictly greater than 1.

# 10. Definition of Subset Neutrosophic Underset

Let $\mathcal{U}$ be a universe of discourse.

***Neutrosophic Underset*** is a set $M_{under}$ from $\mathcal{U}$ that has at least one element (called underelement)

$z\left(t_{M_{under}}, i_{M_{under}}, f_{M_{under}}\right) \in M_{under}$

whose at least one neutrosophic component $t_{M_{under}}, i_{M_{under}}, f_{M_{under}}$ is partially or totally < 0.

For *example*, the following **underelements**:

$$a(-0.6, 0.9, 0.3), b(0, -1.1, [0.8, 0.9]),$$
$$c([0.2, 0.5], \{0.3, 0.7\}, [-0.6, 0.5])$$

since -0.6 < 0 (undertruth, or undermembership), -1.1 < 0 (underindeterminacy), and respectively

$[-0.6, 0.5]$ is partially < 0 (underfalsehood, or undernonmembership).

Therefore, a neutrosophic underset has elements with negative neutrosophic components.

# 11. Definition of Subset Neutrosophic Offset

We now introduce for the first time the Neutrosophic Offset.

Let $\mathcal{U}$ be a universe of discourse and let $O$ be a neutrosophic set in $\mathcal{U}$, i. e.

$$O \subset \mathcal{U}, O = \{x(T_o, I_o, F_o), x \in \mathcal{U}\}, \quad (28)$$

where $T_o$ is the truth-membership,
$I_o$ is the indeterminate-membership,
$F_o$ is the false-membership
of generic element $x$ with respect to the set $O$.

{ There are elements that can be both simultaneously, overelement and underelement. For example: $l(0.1, -0.2, 1.3)$. They are called offelements. }

We say that $O$ is a **Neutrosophic Offset**, if there exists at least one element (called **offelement**)

$$y(T_y, I_y, F_y) \in O, \quad (29)$$

whose at least two of its neutrosophic components are partially or totally off the interval [0, 1], such that one neutrosophic component is below 0, i.e.

$$min\{inf(T_y), inf(I_y), inf(F_y)\} < 0,$$

and the other neutrosophic component is above 1, i.e.

$$max\{sup(T_y), sup(I_y), sup(F_y)\} > 1,$$

where $inf = infimum$ and $sup = suprem$;
or O is a **Neutrosophic Offset** if it has at least one overelement and at least one underelement.

Same definition for the **Neutrosophic Offlogic**, **Neutrosophic Offprobability**, **Neutrosophic Offmeasure** etc.

# 12. Neutrosophic Overprism/Underprism/Offprism

## Neutrosophic Overprism

In the 3D-Cartesian (t, i, f)-system of coordinates, the neutrosophic cube defined on [0, 1]x[0, 1]x[0, 1] is extended to [0, Ω]x[0, Ω]x[0, Ω], where the overlimit Ω > 1.

## Neutrosophic Underprism

Similarly, in the 3D-Cartesian (t, i, f)-system of coordinates, the neutrosophic cube defined on [0, 1]x[0, 1]x[0, 1] is extended to [Ψ, 1]x[Ψ, 1]x[Ψ, 1], where the underlimit Ψ < 0.

## Neutrosophic Offprism

Again, in the 3D-Cartesian (t, i, f)-system of coordinates, the neutrosophic cube defined on [0, 1]x[0, 1]x[0, 1] is extended to [Ψ, Ω]x[Ψ, Ω]x[Ψ, Ω], where the overlimit and underlimit verify the inequalities: Ψ < 0 < 1 < Ω.

## Another Example of Single-Valued Neutrosophic Offset

In this case, at least one neutrosophic component is strictly less than 0, and another one is strictly greater than 1.

As examples, the neutrosophic offset A that contains the neutrosophic offelement:

$y_1(-0.8, -0.2, 1.3)$.

Also, the neutrosophic offset B that contains the neutrosophic overelement and respectively the neutrosophic underelement

$y_2(0.3, 0.4, 1.2)$, and

$y_3(-0.2, 0.7, 0.6)$.

For Hesitant Neutrosophic Offset, Interval Neutrosophic Offset, and the General (Subset) Neutrosophic Offset (i. e. $T_O, I_O, F_O$ are any real subsets), this means that at least one neutrosophic component has a part strictly greater than 1 and another neutrosophic component has a part strictly less than 0.

## Numerical Example of Hesitant Neutrosophic Offset

A neutrosophic set C that contains the below neutrosophic elements:

$y_1(\{0.1, 0.2\}, \{-0.1, 0.3\}, \{0.4, 0.9, 1.4\})$,

$y_2(\{0.6, 0.7\}, \{0.4\}, \{1, 1.2\})$.

## Numerical Example of Interval Neutrosophic Offset

A neutrosophic set D that contains the below neutrosophic elements:

$y_1([0.7, 0.8], [-0.2, 0], [0.0, 0.3])$,

$y_2([0.9, 1.3], [0.5, 0.5], [-0.2, -0.1])$.

# 13. Definition of Non-Standard Neutrosophic Offset

The definition of *Non-Standard Neutrosophic Offset* is an extension of the previous one from standard real subsets to non-standard real subsets $T_O, I_O, F_O$.

This is not used in practical applications, but are defined only from a philosophical point of view, i.e. to make distinction between *absolute* (truth, indeterminacy, falsehood) and *relative* (truth, indeterminacy, falsehood) respectively.

A statement is considered *absolute* if it occurs in all possible worlds, and *relative* if it occurs in at least one world.

Let $\mathcal{U}$ be a universe of discourse, and ${}^-O^+$ be a non-standard neutrosophic set in $\mathcal{U}$, i.e. ${}^-O^+ \subset \mathcal{U}$, and ${}^-O^+ = \{x({}^-T_O{}^+, {}^-I_O{}^+, {}^-F_O{}^+), x \in \mathcal{U}\}$, where ${}^-T_O{}^+, {}^-I_O{}^+, {}^-F_O{}^+$ are non-standard real subsets.

If there exists at least one element

$$z({}^-T_z{}^+, {}^-I_z{}^+, {}^-F_z{}^+) \in {}^-O^+ \quad (30)$$

whose at least one of its non-standard neutrosophic components ${}^-T_z{}^+, {}^-I_z{}^+, {}^-F_z{}^+$ is partially or totally off the non-standard unit interval $]^-0, 1^+[$ , the ${}^-O^+$ is called a ***Non-Standard Neutrosophic Offset***.

Similar definitions for the hyper monads ${}^-O$ and $O^+$ respectively (i.e. sets of hyper-real numbers in non-standard analysis), included into the universe of discourse $\mathcal{U}$, i.e. ${}^-O = \{x({}^-T_O, {}^-I_O, {}^-F_O), x \in \mathcal{U}\}$ and respectively $O^+ = \{x(T_O{}^+, I_O{}^+, F_O{}^+), x \in \mathcal{U}\}$, where ${}^-T_O, {}^-I_O, {}^-F_O$ and respeectively $T_O{}^+, I_O{}^+, F_O{}^+$ are non-standard real subsets.

## Example of Non-Standard Neutrosophic Offset

The neutrosophic set ${}^{-}E^{+}$ that contains the element $w(\,]^{-}0, 1.1^{+}[, \{0.5, 0.6\}, ]^{-}(-0.2), 0.9^{+}[\,)$.

# 14. Neutrosophic Offset

**Neutrosophic Offset** is a set which is both neutrosophic overset and neutrosophic underset. Or, a neutrosophic offset is a set which has some elements such that at least two of their neutrosophic components are one below 0 and the other one above 1.

## Remark

Overtruth means overconfidence.

For example, a set G that contains the following elements:

$$x_1(0.2, \{-0.2, 0.9\}, [0.1,0.5]), x_2([1, 1.5], 0.6, 0.7)$$

is a neutrosophic offset.

# 15. Particular Cases of Neutrosophic Offset

Let also introduce for the first time the notions of **Fuzzy Offset** and respectively of **Intuitionistic Fuzzy Offset**.

(Similar definitions for Fuzzy Overlogic and respectively of Intuitionistic Fuzzy Overlogic.)

## Fuzzy Offset

Let $\mathcal{U}$ be a universe of discourse, and let $O_{fuzy}$ be a fuzzy set in $\mathcal{U}$, i.e. $O_{fuzzy} \subset \mathcal{U}$,

$$O_{fuzzy} = \left\{x\left(T_{O_{fuzzy}}\right), x \in \mathcal{U}\right\}, \qquad (31)$$

where $T_{O_{fuzzy}}$ is the degree of truth-membership of the element $x$ with respect to the fuzzy set $O_{fuzzy}$, where $T_{O_{fuzzy}} \subseteq [0, 1]$.

We say that $O_{fuzzy}$ is a **Fuzzy Offset**, if there exists at least one element $y(T_y) \in O_{fuzzy}$, such that $T_y$ is partially or totally above 1, and another element $z(T_z) \in O_{fuzzy}$ such that $T_z$ is partially or totally below 0.

For example the set G that contains the elements: $y(1.2)$, $z(-0.3)$, $w([-0.1, 0.3])$, $v\big((0.9, 1.1)\big)$.

## Intuitionistic Fuzzy Offset

Let $\mathcal{U}$ be a universe of discourse, and let $O_{intuitionistic}$ be an intuitionistic fuzzy set in $\mathcal{U}$, i.e.

$O_{intuitionistic} \subset \mathcal{U}$,

$O_{intuitionistic} = \{x(T_{O_{intuitionistic}}, F_{O_{intuitionistic}}), x \in \mathcal{U}\}$,

where $T_{O_{intuitionistic}}$ is the degree of truth-membership and $F_{O_{intuitionistic}}$ is the degree of falsehood-nonmembership of the element $x$ with respect to the intuitionistic fuzzy set $O_{intuitionistic}$, where $T_{O_{intuitionistic}}, F_{O_{intuitionistic}} \subseteq [0,1]$.

We say that $O_{intuitionistic}$ is an **Intuitionistic Fuzzy Offset**, if there exists at least one offelement $y\left(T_y, F_y\right)$ such that one of the components $T_y$ or $F_y$ is partially or totally above 1, while the other ine is partially or totally below 0; or there exist at least one overelement and at least one underelement.

For example the set G conatining the below elements:
$y(1.3, 0.9), z(0.2, -0.1)$,
$w([-0.2, 0.2], 0.4)$,
$v\left(0.2, (0.8, 1.1)\right)$.

# 16. Other Particular Cases of Neutrosophic Offset

There are two particular cases of the neutrosophic offset that were presented before:

## Neutrosophic Overset

**The Neutrosophic Overset**, which is a neutrosophic set $O_{over}$ that has at least one element $w(T_w, I_w, F_w) \in O_{over}$ whose at least one of its neutrosophic components $T_w, I_w, F_w$ is partially or totally > 1, and no neutrosophic component of no element is partially or totally < 0.

### Example

$O_{over} = \{w_1\langle 1.2, 0.3, 0.0\rangle, w_2\langle 0.9, 0.1, 0.2\rangle\}$
where there is a neutrosophic component is > 1, and one has no neutrosophic components < 0.

## Neutrosophic Underset

2. **The Neutrosophic Underset**, which is a neutrosophic set $O_{under}$ that has at least one element $z(T_z, I_z, F_z) \in O_{underset}$ whose at least one of its neutrosophic components $T_z, I_z, F_z$ is partially or totally < 0, and no neutrosophic component of no element is partially or totally > 1.

### Example

$$O_{under} = \{z_1\langle 0.2, 0.3, -0.1\rangle, z_2\langle -0.4, 0.0, 0.6\rangle, z_3\langle 0.8, 0.2, 0.3\rangle\}$$

where no neutrosophic component is > 1, and one has neutrosophic components < 0.

## Remark

Similar definitions and examples for Neutrosophic Overlogic, Neutrosophic Overprobability, Neutrosophic Overstatistics, Neutrosophic Overmeasure, etc., respectively for: Neutrosophic Underlogic, Neutrosophic Underprobability, Neutrosophic Understatistics, Neutrosophic Undermeasure etc., that will include both cases.

For simplicity, we will use the notion of Neutrosophic Offset, Neutrosophic Offprobability, Neutrosophic Offstatistics, Neutrosophic Offmeasure etc. that will include both cases.

If one believes that there are neutrosophic components off the classical unitary interval [0, 1], but one not knows if the neutrosophic components are over 1 or under 0, it is better to consider the most general case, i.e. the neutrosophic offset.

As **another example**, an element of the form $x\langle -03, 0.4, 1.2\rangle$ belongs neither to Neutrosophic Overset, nor to a Neutrosophic Underset, but to the general case, i.e. to the Neutrosophic Offset.

## Numerical Example of Subset Neutrosophic Offset

The set H containing the below elements:

$y_1(\{0.1\} \cup [0.3, 0.5], (-0.4, -0.3) \cup [0.0, 0.1], \{0.2, 0.4, 0.7\})$,
$y_2\big([1, 1.5], [0.0, 0.2] \cup \{0.3\}, (0.3, 0.4) \cup (0.5, 0.6)\big)$.

## Why Using the Neutrosophis Offset

The neutrosophic offset, with its associates (neutrosophic offlogic, neutrosophic offmeasure, neutrosophic offprobability, neutrosophic offstatistics etc.) may look counter-intuitive, or shocking, since such things were never done before upon our own knowledge.

How would it be possible, for example, that an element belongs to a set in a strictly more than 100% or in a strictly less than 0%?

In the classical, fuzzy, and intuitionistic fuzzy set an element's membership belongs to (or is included in) the unitary interval [0, 1], in the case of single value (or interval- or subset-value respectively).

Similarly, for the classical, fuzzy, and intuitionistic logic, the truth-value of a proposition belongs to (or is included in) the unitary interval [0, 1], in the case of single value (or interval- or subset-value respectively).

In classical probability, the probability of an event belongs to [0, 1], while in imprecise probability, the probability of an event (being a subset) is included in [0, 1].

Yet, just our everyday life and our real world have such examples that inspired us to introduce the neutrosophic offset / offlogic / offprobability / offmeasure.

## Practical Application of the Neutrosophic Overset (Over-Membership)

Let's consider a given University Alpha. At this university a student is considered a full-time student for a given semester if he or she enrolls in courses that are worth all together 15 credit hours. If the student John enrolls only in 3 credit hours, one says that John's degree of membership (degree of appurtenance) to the University Alpha is $\frac{3}{15} = 0.2 < 1$.

Similarly, if student George enrolls in 12 credit hours, his degree of membership is $\frac{12}{15} = 0.8 < 1$.

Therefore, John and George partially belong to the University of Alpha.

But Mary, who enrolls in 15 credit hours, fully belongs to the University Alpha, since her degree of membership is $\frac{15}{15} = 1$.

Yet, the University Alpha allows students to enroll in more than 15 credit hours, up to 18 credit hours. So, a student can carry an overload. Student Oliver enrolls in 18 credit hours; therefore, his degree of membership is $\frac{18}{15} = 1.2 > 1$.

It is clear that the university has to make distinction, for administrative and financial reasons, between the students who are partially enrolled, totally enrolled, or over loaded (over enrolled).

In general, for a student $x$, one has $x(T, I, F) \in Alpha$, where $0 \leq T, I, F \leq 1.2$, and

$$0 \leq T + I + F \leq 1.2 + 1.2 + 1.2 = 3.6,$$

in the case of a single-valued neutrosophic overset.

# Practical Application of Neutrosophic Overset with Dependent and Independent T, I, F.

Let's take a similar example, with a University Beta, where a full-time student has 15 credit hours, but a student is allowed to enroll in up to 21 credit hours.

If the student Natasha enrolls in 21 credit hours (the maximum allowed), her degree of membership to the University Beta is $\frac{21}{15} = 1.4$.

In general, for a single-valued, neutrosophic overset, a student $y$ has the appurtenance to the University Beta $y(T, I, F) \in Beta$, where $0 \leq T, I, F \leq 1.4$.

a. If the three sources that give information about T, I, and F respectively are independent, then:

$0 \leq T + I + F \leq 1.4 + 1.4 + 1.4 = 4.2$, and one has

$T + I + F = 4.2$ for complete information, and

$T + I + F < 4.2$ for incomplete information.

b. If the three sources are dependent of each other, then $0 \leq T + I + F \leq 1.4$, and one has

$T + I + F = 1.4$ for complete information,

and $T + I + F < 1.4$ for incomplete information.

c. If the two sources are dependent, let's say T and I, while F is independent from them, then:

$0 \leq T + I + F \leq 1.4 + 1.4 = 2.8$, and one has

$T + I + F = 2.8$ for complete information,

and $T + I + F < 2.8$ for incomplete information.

And so on.

## Another Practical Example of Overset (Over-Membership)

A factor worker, Adrian, has the working norm of 40 hours per week as a full-time salary employee.

If he works less than 40 hours, he is paid less money.

Let's say Martin works only 30 hours per week. Then Martin's membership (appurtenance) to this factory is $\frac{30}{40} = 75\% = 0.75$. If he works overtime, he is paid more.

Let's say Angela works 45 hours per week, then her membership is $\frac{45}{40} = 101.25\% = 1.0125 > 1$.

## Practical Example of Offset (Negative-Membership)

Let's consider the Department of Secret Service of country $C$ be:

$DSA_C = \{A_1, A_2, \dots, A_{1000}\}$,

such that each agent $A_j$ ($j \in \{1, 2, \dots, 1000\}$) works full-time for it.

But, among them, there is a double-agent, $A_5$, who spies for the enemy country $E$. The membership degree to $DSA_C$ of, e.g., agent $A_3$ is positive, because he is not a double-agent, but a dedicated worker, while the membership degree to $DSA_C$ of double-agent $A_5$ is negative, since he produces much damage to his country. On the other hand, the degree of membership with respect to country $E$ of double-agent $A_5$ is positive, while the membership degree with respect to country $E$ of agent $A_3$ is negative (under-membership).

Of course, the system of reference counts.

# 17. Definition of Label Neutrosophic Offset

Let's consider a set of labels:

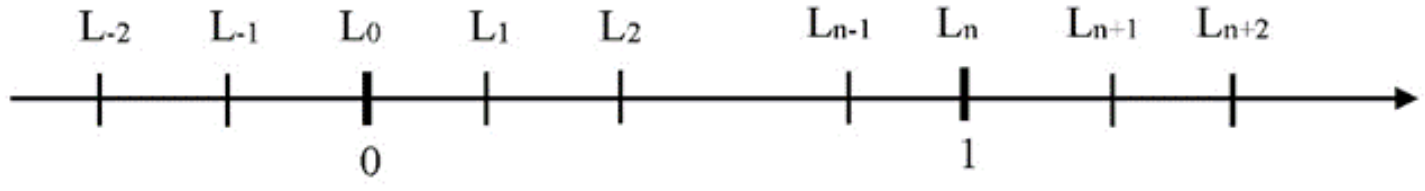

*Fig. 1*

Let's consider $\mathcal{U}$ a universe of discourse and a neutrosophic set $A_L \subset \mathcal{U}$ such that each element $x_L\langle T_L, I_L, F_L\rangle \in A_L$ has all its neutrosophic components

$$T_L, I_L, F_L \subseteq \{L_0, L_1, L_2, \dots, L_{n-1}, L_n\}. \quad (32)$$

This is called a **Label Neutrosophic Set**.

Now, a **Label Neutrosophic Offset** $O_L \subset \mathcal{U}$ is a label neutrosophic set such that it contains some elements that have at least one label component that is strictly greater than $L_n \equiv 1$ and at least one label component that is less than $L_0 \equiv 0$.

Similar definitions for the **Label Neutrosophic Overset** and respectively **Label Neutrosophic Underset**.

# 18. Comment about the Classical Universe of Discourse (Universal Set)

Consulting several dictionaries about this definition, we observed that it is too general.

In the *dictionary.com*[1], the "universal set" in mathematics is "the set of all elements under discussion for a given problem", and "universe of discourse" in logic is "the aggregate of all the objects, attributes, and relations assumed or implied in a given discussion".

In the *Webster-Merriam Dictionary*[2], the "universe of discourse" is "an inclusive class of entities that is tacitly implied or explicitly delineated as the subject of a statement, discourse, or theory".

In the *HarperCollins Dictionary of Mathematics* (1991), it is "some specific class large enough to include all the elements of any set relevant to the subject matter".

---

1 Dictionary.com, http://www.dictionary.com/browse/universe--of--discourse.

2 Merriam-Webster Dictionary, http://www.merriam-webster.com/dictionary/universe%20of%20discourse.

# 19. (Counter-)Example to the Universal Set

Let's see the following counter-example.

One considers the set of integers $\mathbb{Z}$ as the universe of discourse.

$M = \{3, 4\}$ and $P = \{5, 6\}$ are two subsets of the universal set. If we compute

$M + P = \{3 + 5, 3 + 6, 4 + 5, 4 + 6\} = \{8, 9, 10\}$,

then the result is in $\mathbb{Z}$.

But, calculating

$\frac{M}{P} = \left\{\frac{3}{5}, \frac{3}{6}, \frac{4}{5}, \frac{4}{6}\right\} \notin \mathbb{Z}$.

Now, a question arises: Is $\mathbb{Z}$ a universal set of $M$ and $P$, or not? If we do only additions, the answer is yes, if we do divisions, the answer may be no.

That's why, in our opinion, the exact definition of the Universe of Discourse (or Universal Set) should be: a larger class that includes all sets involved in the matter, together with all resulted sets after all their aggregations.

In other words, the universal set's structure should be specified if one applies operators on its subsets.

# 20. Neutrosophic Universe of Discourse (Neutrosophic Universal Set)

In the classical Universe of Discourse, $\mathcal{U}$, all elements that belong to it, $x \in \mathcal{U}$, have the understandable neutrosophic truth-value $x(1, 0, 0)$, i.e. they are totally included in $\mathcal{U}$.

We extend now, for the first time, the classical universe of discourse to the Neutrosophic Universe of Discourse, $\mathcal{U}_N$, which means that all elements belonging to $\mathcal{U}_N$ have the neutrosophic truth-value $x(T_{\mathcal{U}_N}, I_{\mathcal{U}_N}, F_{\mathcal{U}_N})$ where $T_{\mathcal{U}_N}, I_{\mathcal{U}_N}, F_{\mathcal{U}_N}$ are, in general, subsets of [0, 1].

Also, if $A$ and $B$ are subsets of $\mathcal{U}_N$, then $A * B$ should also be a subset of $\mathcal{U}_N$, where "$*$" is any operation defined into the problem to solve.

A neutrosophic set is a set $A \subset \mathcal{U}_N$ of the form

$$A = \{\langle x, T_A, I_A, F_A\rangle, x \in \mathcal{U}_N, \text{and } T_A \leq T_{\mathcal{U}}, I_A \geq I_{\mathcal{U}}, F_A \geq F_{\mathcal{U}}\}. \tag{33}$$

In other words, "$A \subset \mathcal{U}_N$" is just the neutrosophic inclusion for crisp neutrosophic components.

Surely, there are other ways to define the neutrosophic inclusion, for example $T_A \leq T_{\mathcal{U}}$, $I_A \leq I_{\mathcal{U}}$, $F_A \geq F_{\mathcal{U}}$, and $T_{\mathcal{U}}, I_{\mathcal{U}}, F_{\mathcal{U}}$ are crisp numbers in the ujnitary interval [0, 1], the three above inequlities among the neutrosophic components are subsets, then:

$T_A \leq T_{\mathcal{U}}$ will mean:

$$inf(T_A) \leq inf(T_{\mathcal{U}})$$
$$sup(T_A) \leq sup(T_{\mathcal{U}})$$

while $I_A \geq I_{\mathcal{U}}$ will mean:

$$inf(I_A) \geq inf(I_{\mathcal{U}})$$

$$sup(I_A) \geq sup(I_{\mathcal{U}})$$

and similarly $F_A \geq F_{\mathcal{U}}$ will mean:

$$inf(F_A) \geq inf(F_{\mathcal{U}})$$
$$sup(F_A) \geq sup(F_{\mathcal{U}}).$$

## Numerical Example of Neutrosophic Universe

$\mathcal{U}_N = \{\langle x_1; 0.8, 0.2, 0.1\rangle, \langle x_2; 0.3, 0.6, 0.7\rangle, \langle x_3; 1, 0, 0\rangle\}$.

And a neutrosophic set included in it:

$A = \{\langle x_1; 0.7, 0.3, 0.4\rangle, \langle x_2; 0.3, 0.6, 0.8\rangle\}$

(No neutrosophic operation defined.)

## Practical Example of Neutrosophic Universe

All members of an association, such that some of them partially belong to and rarely are involved into association affairs, others totally belong, while about a third category of members is unclear their appurtenance or non-appurtenance to the association. (No neutrosophic aggregation was specified.)

## Neutrosophic Applications

For our needs in engineering, cybernetics, military, medical and social science applications, where we mostly use the following operations:

- neutrosophic complement/negation
- neutrosophic intersection / AND
- neutrosophic union / OR,

while other operations (neutrosophic implication, neutrosophic inclusion, neutrosophic strong / weak disjunctions, neutrosophic equivalence, etc.) are composed from the previous three, a Neutrosophic Universal Set is a set

$$\mathcal{U}_N = \left(\langle x, T_{\mathcal{U}_N}, \mathrm{I}_{\mathcal{U}_N}, F_{\mathcal{U}_N}\rangle, \cup, \cap, \mathcal{C}\right) \quad (34)$$

closed under neutrosophic union, neutrosophic intersection, and neutrosophic complement, such that $\mathcal{U}_N$ includes all elements of the sets involved into the problem to solve.

Therefore, $\mathcal{U}_N$ is a **Neutrosophic Universal Boolean Algebra**.

Consequently, the **Neutrosophic Offuniverse of Discourse** (or **Neutrosophic Offuniversal Set**), $\mathcal{U}_O$, means a neutrosophic universe of discourse such that all elements that belong to $\mathcal{U}_O$ have the neutrosophic offtruth value $x(T_{\mathcal{U}_O}, \mathrm{I}_{\mathcal{U}_O}, F_{\mathcal{U}_O})$ and there exist some elements in $\mathcal{U}_O$ having at least one neutrosophic component partially or totally over 1, and another neutrosophic component partially or totally below 0.

Similarly as for the neutrosophic universal set, if elements of the $A$ and $B$ are subsets of $\mathcal{U}_O$, then $A * B$ should also be a subset of $\mathcal{U}_O$, where $*$ is any operation defined into the problem to solve.

And for applications, a **Neutrosophic Offuniversal Set** is a set

$$\mathcal{U}_{O_N} = \left(\langle x, T_{\mathcal{U}_O}, \mathrm{I}_{\mathcal{U}_O}, F_{\mathcal{U}_O}\rangle, \cup, \cap, \mathcal{C}\right) \quad (35)$$

closed under neutrosophic union, neutrosophic intersection, and neutrosophic complement, such that $\mathcal{U}_O$ includes all elements of the sets involved in the problem to solve, and there exist some elements in $\mathcal{U}_O$ having at least one neutrosophic component partially or totally over 1, and another neutrosophic component partially or totally below 0.

Therefore, $\mathcal{U}_O$ is a **Neutrosophic Offuniversal Boolean Algebra**.

Similar definition for the **Neutrosophic Overuniversal Boolean Algebra** and respectively the **Neutrosophic Underuniversal Boolean Algebra**.

## Numerical Example of Neutrosophic Offuniverse

$\mathcal{U}_O = \{\langle x_1; 1.2, 0.1, 0.3\rangle, \langle x_2; 0.6, 0.7, -0.1\rangle\}$,
and an example of a neutrosophic set $B \subset \mathcal{U}_O$,
$B = \{\langle x_1; 1.0, 0.2, 0.4\rangle, \langle x_2; 0.4, 0.7, 0.0\rangle\}$,
then an example of a neutrosophic offset $C_O \subset \mathcal{U}_O$,
$C_O = \{\langle x_1; 1.1, 0.3, 0.3\rangle, \langle x_2; 0.6, 0.8, -0.1\rangle\}$.
(No neutrosophic operation was specified.)

## Practical Example of Neutrosophic Overuniverse

All students enrolled in, let's say, University Alpha, such that there exists some student which are overloaded. (No neutrosophic aggregation was specified.)

# 21. Neutrosophic Offuniverse (and consequently Neutrosophic Offset)

Let's suppose one has a non-empty set $O$, whose elements are characterized by an attribute "a".

For the attribute "a", there exists a corresponding set $V_a$ of all attribute's values.

The attribute's values can be numerical or linguistic, and they may be discrete, continuous, or mixed.

The set $V_a$ is endowed with a total order $\underset{a}{<}$ (less important than, or smaller than). Consequently, one has $\underset{a}{\leq}$ (that means less important than or equal to, or smaller than and equal to). And the reverse of $\underset{a}{<}$ is $\underset{a}{>}$ (more important than, or greater than). Similarly, the reverse of $\underset{a}{\leq}$ is $\underset{a}{\geq}$ (more important than or equal to, greater than or equal to).

Therefore, for any two elements $v_1$ and $v_2$ from $V_a$, one has: either $v_1 \underset{a}{<} v_2$, or $v_1 \underset{a}{>} v_2$.

Let's define, with respect to this attribute, the following functions:

1. The **Truth-Value Function**:

$$t: V_a \to \mathbb{R}$$

which is a strictly increasing function, i.e. if $v_1 < v_2$, $t(v_1) < t(v_2)$.

Let's suppose there exists a **lower threshold truth** $\tau_T^L \in V_a$ such that $t(\tau_T^L) = 0$, and an **upper threshold truth** $\tau_T^U \in V_a$ such that $(\tau_T^U) = 1$.

If there exists an element $\eta_T^L \in V_a$ such that $\eta_T^L < \tau_T^L$, then $t(\eta_T^L) < t(\tau_T^L) = 0$, therefore one gets a **negative truth-value** [**undertruth**].

Similarly, if there exists an element $\eta_T^U \in V_a$ such that $\eta_T^U > \tau_T^U$, then $t(\eta_T^U) > \mathrm{t}(\tau_T^U) = 1$, therefore one gets an over 1 truth-value [**overtruth**].

2. Analogously, one defines the **Indeterminate-Value Function**:

$$i: V_a \to \mathbb{R}$$

which is also a strictly increasing function, for $v_1 < v_2$ one has $i(v_1) < i(v_2)$ for all $v_1, v_2 \in V_a$.

One supposes there exists a **lower threshold indeterminacy** $\tau_I^L \in V_a$, such that $i(\tau_I^L) = 0$, and an **upper threshold indeterminacy** $\tau_I^U \in V_a$, such that $i(\tau_I^U) = 1$.

If there exists an element $\eta_I^L \in V_a$ such that $\eta_I^L < \tau_I^L$, then $i(\eta_I^L) < i(\tau_I^L) = 0$ , therefore one gets a **negative indeterminate-value** [**underindeterminacy**].

Similarly, if there exists an element $\eta_I^U \in V_a$ such that $\eta_I^U > \tau_I^U$ , then $i(\eta_I^U) > i(\tau_I^U) = 1$, therefore one gets an **over 1 indeterminate-value** [**overindeterminacy**].

3. Eventually, one defines the **False-Value Function**:

$$f: V_a \to \mathbb{R}$$

also a strictly increasing function: for $v_1 < v_2$ one has $f(v_1) < f(v_2)$ for all $v_1, v_2 \in V_a$.

Again, one supposes there exists a **lower threshold falsity** $\tau_F^L \in V_a$ , such that $f(\tau_F^L) = 0$ , and an **upper threshold falsity** $\tau_F^U \in V_a$, such that $f(\tau_F^U) = 1$.

Now, if there exists an element $\eta_F^L \in V_a$ such that $\eta_F^L < \tau_F^L$, then $f(\eta_F^L) < f(\tau_F^L) = 0$, therefore one gets a **negative false-value** [**underfalsity**].

Similarly, if there exists an element $\eta_F^U \in V_a$ such that $\eta_F^U > \tau_F^U$, then $f(\eta_F^U) > f(\tau_F^U) = 1$, therefore one gets an **over 1 false-value** [**overfalsity**].

## Question 1

How big can be the overlimits of $T$, $I$, and $F$ respectively above 1?

*Answer:* It depends on each particular problem or application. It may be subjective, as in the previous two examples with universities, where the overlimits of T, I, F were 1.2 for the University Alpha, and respectively 1.4 for the University Beta. Or it may be objective.

## Notations 1

We denote by

$\Omega_T$ the overlimit of $t$,

$\Omega_I$ the overlimit of $i$,

$\Omega_F$ the overlimit of $f$.

## Remark 3

The overlimits $\Omega_T$, $\Omega_I$, $\Omega_F$ need not be equal. It depends on each particular problem or application too.

## Question 2

How low can be the underlimits of $T$, $I$, and $F$ respectively below 0?

*Same answer:* It depends on each particular problem or application. It may be subjective or objective.

# Notations 2

We denote by
$\Psi_T$ the underlimit of $t$,
$\Psi_I$ the underlimit of $i$,
$\Psi_F$ the underlimit of $f$.

In many cases, the underlimits of the neutrosophic components are equal, i.e.

$\Psi_T = \Psi_I = \Psi_F$

and similarly for the overlimits, i.e.

$\Omega_T = \Omega_I = \Omega_F$

but there also are cases and applications when these two above double equalities do not hold.

## 22. Inequalities

The truth-value function:

$$t(v) = \frac{v-\tau_T^L}{\tau_T^U-\tau_T^L}\text{, thus } \frac{\Psi_T-\tau_T^L}{\tau_T^U-\tau_T^L} \le t(v) \le \frac{\Omega_T-\tau_T^L}{\tau_T^U-\tau_T^L}. \quad (36)$$

The indeterminate-value function:

$$i(v) = \frac{v-\tau_I^L}{\tau_I^U-\tau_I^L}\text{, thus } \frac{\Psi_I-\tau_I^L}{\tau_I^U-\tau_I^L} \le i(v) \le \frac{\Omega_F-\tau_F^L}{\tau_I^U-\tau_I^L}. \quad (37)$$

The falsehood-value function:

$$f(v) = \frac{v-\tau_F^L}{\tau_F^U-\tau_F^L}\text{, thus } \frac{\Psi_F-\tau_F^L}{\tau_F^U-\tau_F^L} \le f(v) \le \frac{\Omega_F-\tau_F^L}{\tau_F^U-\tau_F^L}. \quad (38)$$

# 23. The Single-Valued Triangular Neutrosophic Offnumber

Let $\bar{a} = <(a_1, a_2, a_3); w_{\bar{a}}, u_{\bar{a}}, y_{\bar{a}}>$, where $a_1$, $a_2$, $a_3$ are real numbers and $a_1 \le a_2 \le a_3$,
with $w_{\bar{a}} \in [\Psi_T, \Omega_T]$, $u_{\bar{a}} \in [\Psi_I, \Omega_I]$, $y_{\bar{a}} \in [\Psi_F, \Omega_F]$,
and also $\Psi_T < 0 < 1 < \Omega_T$, and $\Psi_I < 0 < 1 < \Omega_I$, and $\Psi_F < 0 < 1 < \Omega_F$, whose truth-membership function $T_{\bar{a}}(x)$, indeterminacy-membership function $I_{\bar{a}}(x)$, and respectively falsity-membership function $F_{\bar{a}}(x)$ are:

$$T_{\bar{a}}(x) = \begin{cases} (x-a_1)w_{\bar{a}} / (a_2 - a_1), if\,(a_1 \le x < a_2); \\ w_{\bar{a}}, if\,(x = a_2); \\ (a_3 - x)w_{\bar{a}} / (a_3 - a_2), if\,(a_2 < x \le a_3); \\ \Psi_T, otherwise. \end{cases}, \quad (39)$$

$$I_{\bar{a}}(x) = \begin{cases} [a_2 - x + u_{\bar{a}}(x - a_1)] / (a_2 - a_1), if\,(a_1 \le x < a_2); \\ u_{\bar{a}}, if\,(x = a_2); \\ [x - a_2 + u_{\bar{a}}(a_3 - x)] / (a_3 - a_2), if\,(a_2 < x \le a_3); \\ \Omega_I, otherwise. \end{cases},$$

(40)

$$F_{\bar{a}}(x) = \begin{cases} [a_2 - x + y_{\bar{a}}(x - a_1)] / (a_2 - a_1), if\,(a_1 \le x < a_2); \\ y_{\bar{a}}, if\,(x = a_2); \\ [x - a_2 + y_{\bar{a}}(a_3 - x)] / (a_3 - a_2), if\,(a_2 < x \le a_3); \\ \Omega_F, otherwise. \end{cases}$$

(41)

Then ā is called a single-valued triangular offnumber.

It should be observed that it is defined similar to the single-valued neutrosophic triangular number, with the distinctions that “0” was replaced by corresponding “Ψ” for each neutrosophic component, while “1” was replaced by the corresponding “Ω” for each neutrosophic component.

Also, of course, $w_{ā}$, $u_{ā}$, and $y_{ā}$ may be > 1 or < 0.

# 24. The Single-Valued Trapezoidal Neutrosophic Offnumber

Let ā = <($a_1$, $a_2$, $a_3$, $a_4$); T(a), I(a), F(a)>, where $a_1$, $a_2$, $a_3$, $a_4$ are real numbers and $a_1 \le a_2 \le a_3 \le a_4$, with

$w_{\bar{a}} \in [\Psi_T, \Omega_T]$, $u_{\bar{a}} \in [\Psi_I, \Omega_I]$, $y_{\bar{a}} \in [\Psi_F, \Omega_F]$,

where $\Psi_T < 0 < 1 < \Omega_T$, and $\Psi_I < 0 < 1 < \Omega_I$, and $\Psi_F < 0 < 1 < \Omega_F$, whose truth-membership function $T_{\bar{a}}(x)$, indeterminacy-membership function $I_{\bar{a}}(x)$, and respectively falsity-membership function $F_{\bar{a}}(x)$ are:

$$T_{\bar{a}}(x) = \begin{cases} (x-a_1)w_{\bar{a}} / (a_2-a_1), if\,(a_1 \le x < a_2); \\ w_{\bar{a}}, if\,(a_2 \le x \le a_3); \\ (a_4-x)w_{\bar{a}} / (a_4-a_3), if\,(a_3 < x \le a_4); \\ \Psi_T, otherwise. \end{cases}, \quad (42)$$

$$I_{\bar{a}}(x) = \begin{cases} [a_2-x+u_{\bar{a}}(x-a_1)] / (a_2-a_1), if\,(a_1 \le x < a_2); \\ u_{\bar{a}}, if\,(a_2 \le x \le a_3); \\ [x-a_3+u_{\bar{a}}(a_4-x)] / (a_4-a_3), if\,(a_3 < x \le a_4); \\ \Omega_I, otherwise. \end{cases}$$

(43)

$$F_{\bar{a}}(x) = \begin{cases} [a_2-x+y_{\bar{a}}(x-a_1)] / (a_2-a_1), if\,(a_1 \le x < a_2); \\ y_{\bar{a}}, if\,(a_2 \le x \le a_3); \\ [x-a_3+y_{\bar{a}}(a_4-x)] / (a_4-a_3), if\,(a_3 < x \le a_4); \\ \Omega_F, otherwise. \end{cases}$$

(44)

Then ā is called a single-valued trapezoidal offnumber.

It should be observed that it is defined similar to the single-valued neutrosophic trapezoidal number, with the distinctions that "0" was replaced by corresponding "Ψ" for each neutrosophic component, while "1" was replaced by the corresponding "Ω" for each neutrosophic component.

Also, of course, $w_{ā}$, $u_{ā}$, and $y_{ā}$ may be $> 1$ or $< 0$.

# 25. Degree of Dependence and Independence of the (Sub)Components of Fuzzy Set and Neutrosophic Set

## Refined Neutrosophic Set

We start with the most general definition, that of a *n-valued refined neutrosophic set* $A$. An element $x$ from $A$ belongs to the set in the following way:

$$x(T_1, T_2, \dots, T_p;\ I_1, I_2, \dots, I_r;\ F_1, F_2, \dots, F_s) \in A, \qquad (45)$$

where $p, r, s \geq 1$ are integers, and $p + r + s = n \geq 3$, where

$$T_1, T_2, \dots, T_p;\ I_1, I_2, \dots, I_r;\ F_1, F_2, \dots, F_s \qquad (46)$$

are respectively sub-membership degrees, sub-indeterminacy degrees, and sub-nonmembership degrees of element $x$ with respect to the $n$-valued refined neutrosophic set $A$. Therefore, one has $n$ (sub)components.

Let's consider all of them being crisp numbers in the interval $[0, 1]$.

## General case

Now, in general, let's consider $n$ crisp-components (variables):

$$y_1, y_2, \dots, y_n \in [0, 1]. \qquad (47)$$

If all of them are 100% independent two by two, then their sum:

$$0 \leq y_1 + y_2 + \dots + y_n \leq n. \qquad (48)$$

But if all of them are 100% dependent (totally interconnected), then

$$0 \leq y_1 + y_2 + \dots + y_n \leq 1. \qquad (49)$$

When some of them are partially dependent and partially independent, then

$y_1 + y_2 + \ ... + y_n \in (1, n).$ (50)

For example, if $y_1$ and $y_2$ are 100% dependent, then

$0 \leq y_1 + y_2 \leq 1,$ (51)

while other variables $y_3, \dots, y_n$ are 100% independent of each other and also with respect to $y_1$ and $y_2$, then

$0 \leq y_3 + \cdots + y_n \leq n - 2,$ (52)

thus

$0 \leq y_1 + y_2 + y_3 + \cdots + y_n \leq n - 1.$ (53)

## Fuzzy Set

Let $T$ and $F$ be the membership and respectively the nonmembership of an element $x(T, F)$ with respect to a fuzzy set $A$, where $T, F$ are crisp numbers in $[0, 1]$.

If $T$ and $F$ are 100% dependent of each other, then one has as in classical fuzzy set theory

$0 \leq T + F \leq 1.$ (54)

But if $T$ and $F$ are 100% independent of each other (that we define now for the first time in the domain of fuzzy setand logic), then

$0 \leq T + F \leq 2.$ (55)

We consider that the sum $T + F = 1$ if the information about the components is complete, and $T + F < 1$ if the information about the components is incomplete.

Similarly, $T + F = 2$ for complete information, and $T + F < 2$ for incomplete information.

For complete information on T and F, one has $T + F \in [1, 2]$.

# 26. Degree of Dependence and Degree of Independence for two Components

In general (see [1], 2006, pp. 91-92), the sum of two components $x$ and $y$ that vary in the unitary interval [0, 1] is:

$$0 \le x + y \le 2 - d^\circ(x, y), \quad (56)$$

where $d^\circ(x, y)$ is the *degree of dependence* between $x$ and $y$.

Therefore $2 - d^\circ(x, y)$ is the *degree of independence* between $x$ and $y$.

Of course, $d^\circ(x, y) \in [0, 1]$, and it is zero when $x$ and $y$ are 100% independent, and 1 when $x$ and $y$ are 100% dependent.

In general, if T and F are $d$% dependent [and consequently $(100 - d)$% independent], then

$$0 \le T + F \le 2 - d/100. \quad (57)$$

## Example of Fuzzy Set with Partially Dependent and Partially Independent Components

As an example, if $T$ and $F$ are 75% (= 0.75) dependent, then

$$0 \le T + F \le 2 - 0.75 = 1.25. \quad (58)$$

## Neutrosophic Set

Neutrosophic set is a general framework for unification of many existing sets, such as fuzzy set (especially intuitionistic fuzzy set), paraconsistent set, intuitionistic set, etc. The main idea of NS is to characterize each value statement in a 3D-Neutrosophic Space, where each dimension of the space represents respectively the membership/truth (T), the nonmembership/falsehood (F), and the indeterminacy with respect to membership/nonmembership (I) of the statement under consideration, where T, I, F are standard or non-standard real subsets of $]^{-}0, 1^{+}[$ with not necessarily any connection between them.

For software engineering proposals the classical unit interval [0, 1] is used.

For single valued neutrosophic set, the sum of the components (T+I+F) is (see [1], p. 91):

$0 \leq T+I+F \leq 3,$ (59)

when all three components are independent;

$0 \leq T+I+F \leq 2,$ (60)

when two components are dependent, while the third one is independent from them;

$0 \leq T+I+F \leq 1,$ (61)

when all three components are dependent.

When three or two of the components T, I, F are independent, one leaves room for incomplete information (sum < 1), paraconsistent and contradictory information (sum > 1), or complete information (sum = 1).

If all three components T, I, F are dependent, then similarly one leaves room for incomplete information (sum < 1), or complete information (sum = 1).

The dependent components are tied together.

Three sources that provide information on T, I, and F respectively are independent if they do not communicate with each other and do not influence each other.

Therefore, max{T+I+F} is in between 1 (when the degree of independence is zero) and 3 (when the degree of independence is 1).

## Examples of Neutrosophic Set with Partially Dependent and Partially Independent Components

The max{T+I+F} may also get any value in (1, 3).

a) For example, suppose that T and F are 30% dependent and 70% independent (hence $T + F \leq 2-0.3 = 1.7$), while I and F are 60% dependent and 40% independent (hence $I + F \leq 2-0.6 = 1.4$). Then $\max\{T + I + F\} = 2.4$ and occurs for $T = 1$, $I = 0.7$, $F = 0.7$.

b) Second example: suppose T and I are 100% dependent, but I and F are 100% independent. Therefore, $T + I \leq 1$ and $I + F \leq 2$, then $T + I + F \leq 2$.

## More on Refined Neutrosophic Set

The Refined Neutrosophic Set [4], we introduced for the first time in 2013. In this set the neutrosophic component (T) is split into the subcomponents ($T_1, T_2, \ldots, T_p$) which represent types of truths (or sub-truths), the neutrosophic component (I) is split into the subcomponents ($I_1, I_2, \ldots, I_r$) which represents types of indeterminacies (or sub-indeterminacies), and the neutrosophic components (F) is split into the subcomponents ($F_1, F_2, \ldots, F_s$) which represent types of falsehoods (or sub-falsehoods), such that p, r, s are integers $\geq 1$ and $p + r + s = n \geq 4$. (62)

When n = 3, one gets the non-refined neutrosophic set. All $T_j$, $I_k$, and $F_l$ subcomponents are subsets of [0, 1].

Let's consider the case of refined single-valued neutrosophic set, i.e. when all n subcomponents are crisp numbers in [0, 1].

Let the sum of all subcomponents be:

$$S = \sum_{1}^{p} T_j + \sum_{1}^{r} I_k + \sum_{1}^{s} F_l \qquad (63)$$

When all subcomponents are independent two by two, then

$$0 \leq S \leq n. \qquad (64)$$

If *m* subcomponents are 100% dependent, $2 \leq m \leq n$, no matter if they are among $T_j$, $I_k$, $F_l$ or mixed, then

$$0 \leq S \leq n - m + 1 \qquad (65)$$

and one has $S = n - m + 1$ when the information is complete, while $S < n - m + 1$ when the information is incomplete.

## Examples of Refined Neutrosophic Set with Partially Dependent and Partially Independent Components

Suppose T is split into $T_1$, $T_2$, $T_3$, and I is not split, while F is split into $F_1$, $F_2$. Hence one has:

$$\{T_1, T_2, T_3; I; F_1, F_2\}. \qquad (66)$$

Therefore, a total of 6 (sub)components.

a) If all 6 components are 100% independent two by two, then:

$$0 \leq T_1 + T_2 + T_3 + I + F_1 + F_2 \leq 6 \qquad (67)$$

b) Suppose the subcomponets $T_1$, $T_2$, and $F_1$ are 100% dependent all together, while the others

are totally independent two by two and independent from $T_1$, $T_2$, $F_1$, therefore:

$0 \leq T_1 + T_2 + F_1 \leq 1$ (68)

whence

$0 \leq T_1 + T_2 + T_3 + I + F_1 + F_2 \leq 6 - 3 + 1 = 4.$ (69)

One gets equality to 4 when the information is complete, or strictly less than 4 when the information is incomplete.

c) Suppose in another case that $T_1$ and I are 20% dependent, or $d°(T_1, I) = 20\%$, while the others similarly totally independent two by two and independent from $T_1$ and I, hence

$0 \leq T_1 + I \leq 2 - 0.2 = 1.8$ (70)

whence

$0 \leq T_1 + T_2 + T_3 + I + F_1 + F_2 \leq 1.8 + 4 = 5.8,$ (71)

since $0 \leq T_2 + T_3 + F_1 + F_2 \leq 4.$ (72)

Similarly, to the right one has equality for complete information, and strict inequality for incomplete information.

## More on the Degree of Dependence and Independence of the Neutrosophic Set

For the neutrosophic set, one has

$0 \leq t + i + f \leq 1$ (73)

for $d^o(t, i, f) = 100\%$ {degree of dependence between the neutrosophic components t, i, f)};

$0 \leq t + i + f \leq 3$ (74)

for $d^o(t, i, f) = 0\%$.

1. Therefore, in the general case, when the degree of dependence of all three components together is $d^o(t,i,f) \in [0,1]$, and $t,i,f \in [0,1]$, then:

$$0 \leq t+i+f \leq 3-2 \cdot d^o(t,i,f). \qquad (75)$$

If the degrees of dependence between two by two components is as follows:

$$d^o(t,i) \in [0,1],$$
$$d^o(i,f) \in [0,1],$$
$$d^o(f,t) \in [0,1], \qquad (76)$$

then one has respectively:

$$0 \leq t+i \leq 2-d^o(t,i) \in [1,2],$$
$$0 \leq i+f \leq 2-d^o(i,f) \in [1,2],$$
$$0 \leq f+t \leq 2-d^o(f,t) \in [1,2], \qquad (77)$$

whence:

$$0 \leq t+i+f \leq max\{2-d^o(t,i), 2-d^o(i,f), 2-d^o(f,t)\}+1 = 2-min\{d^o(t,i), d^o(i,f), d^o(f,t)\}+1 = 3-min\{d^o(t,i), d^o(i,f), d^o(f,t)\}. \qquad (78)$$

Therefore:

$$0 \leq t+i+f \leq 3-min\{d^o(t,i), d^o(i,f), d^o(f,t)\}. \qquad (79)$$

# 27. Degree of Dependence and Independence of Neutrosophic Offcomponents

Let's suppose one has:

$$t_l \leq t \leq t_u$$
$$i_l \leq i \leq i_u$$
$$f_l \leq f \leq f_u$$

where

$t_l$ = lowest value of $t$;

$t_u$ = highest (upper) value of $t$;

$i_l$ = lowest value of $i$;

$i_u$ = highest (upper) value of $i$;

$f_l$ = lowest value of $u$;

$f_u$ = highest (upper) value of $u$.

1. If all three sources providing information on $t, i, f$ respectively are independent two by two, then

$$t_l + i_l + f_l \leq t + i + f \leq t_u + i_u + f_u. \quad (80)$$

2. If all three sources providing information on $t, i, f$ respectively are dependent, then

$$min\{t_l + i_l + f_l\} \leq t + i + f \leq max\{t_u + i_u + f_u\}. \quad (81)$$

3. If two sources, let suppose those providing information on $t$ and $i$ are dependent, then:

$$min\{t_l, i_l\} \leq t + i \leq max\{t_u, i_u\}, \quad (82)$$

and the third source, providing information on f is independent from both of them, then:

$$f_l \leq f \leq f_u. \quad (83)$$

Therefore:

$$min\{t_l, i_l\} + f_l \leq t + i + f \leq max\{t_u, i_u\} + f_u. \quad (84)$$

Similarly, if $t$ and $f$ are dependent, and $i$ independent from them:

$min\{t_l, f_l\} + i_l \leq t + i + f \leq max\{t_u, f_u\} + i_u$. (85)

Or, if i and f are dependent, and t is independent from them:

$min\{i_l, f_l\} + t_l \leq t + i + f \leq max\{i_u, f_u\} + t_u$. (86)

4. If the degree of dependence of all three neutrosophic offsources together is any $d^0(t, i, f) \in [0, 1]$, then:

$t_i + i_j + f_l - (t_l + i_l + f_l - min\{t_l, i_l, f_l\}) \cdot d^0(t, i, f)$
$\leq t + i + f \leq t_u + i_u + f_u - (t_u + i_u + f_u -$
$max\{t_u, i_u, f_u\}) \cdot d^0(t, i, f)$. (87)

The first side of this double inequality shows how from the degree of dependence $d^0(t, i, f) = 0$ and corresponding value $t_l + i_l + f_l$ one gradually gets for the degree of dependence $d^0(t, i, f) = 1$ to the value $min\{t_l, i_l, f_l\}$ according to the above inequalities (80) and (81).

Similarly, for the third side of this double inequality: from $d^0(t, i, f) = 0$ and corresponding value $t_u + i_u + f_u$ one gradually gets for $d^0(t, i, f) = 1$ to the value $max\{t_u, i_u, f_u\}$.

5. Let's now suppose the degree of dependence between two neutrosophic offsources as follows:

$d^0(t, i) \in [0, 1]$,
$d^0(i, f) \in [0, 1]$,
$d^0(f, t) \in [0, 1]$. (88)

Then one gets:

a. One has:

$t_l + i_l - (t_l + i_l - min\{t_l, i_l\})d^o(t, i) \leq t + i \leq t_u + i_u - (t_u + i_u - max\{t_u, i_u\}) \cdot d^o(t, i),$ (89)

since for the degree of dependence $d^o(t, i) = 0$ one has

$t_l + i_l \leq t + i \leq t_u + i_u$ (90)

and for the degree of dependence $d^o(t, i) = 1$ one has

$min\{t_l, i_l\} \leq t + i \leq max\{t_u, i_u\}.$ (91)

b. Similarly:

$i_l + f_l - (i_l + f_l - min\{i_l, f_l\}) \cdot d^o(i, f) \leq i + f \leq i_i + f_u - (i_i + f_u - max\{i_i, f_u\})d^o(i, f).$ (92)

c. And:

$f_l + t_l - (f_l + t_l - min\{f_l, t_l\})d^0(f, t) \leq f + t \leq f_u + t_u - (f_u + t_u - max\{f_u, t_u\})d^0(f, t).$ (93)

## Practical Example of Neutrosophic Offset

The company "Inventica" produces electronic devices.

The norm for a full-time worker is 20 electronic devices per week.

The company's policy is the following:

- for each electronic device constructed correctly, the employee gets 1 point (at 20 points the employee gets a full-salary);
- for an electronic device not constructed, the employee gets no points;
- for each electronic device constructed wrongly, the employee loses 2 points (1 point for the wasted material, and 1 point for the labor/time used in building a wrong device);
- the employee also loses points for other damages done to the company;

- the employee gains points for other benefits (besides electronic devices) brought to the company.

The neutrosophic overset is:

$O$ = {all *Inventica* employees}.

The attribute "$a$" = working ($w$).

The set of all attribute's values is

$V_a = V_w = [b, c]$, with $b \leq -40$ and $c \geq 20$,

which is numerical and continuous.

We also considered the case when an electronic device was not finished at the end of the week, so only a part of it done. Otherwise we'd take a discrete set.

Therefore, the minimum underlimit is $\leq -40$, i.e. in the situation when a worker produces wrong electronic devices, but the damage can be done even at a higher proportion (destroying tools, etc.).

In the history of the company, the worst damage has been done two years ago by Jack (-45) who has produced defected electronic devices and destroyed several tools.

The maximum overlimit is > 20, for employees working faster, or doing overtime.

One studies record in the history of the company.

Suppose an employee, Thom, has produced 30 electronic devices last year in the first week of February.

We readjust the set of attribute's values:

$V_w = [b, c]$, with $b \leq -45$ and $c \geq 30$.

The truth-value function, the indeterminate-value function, and the false-value function are, respectively:

$t: V_w \to \mathbb{R}$,

$i: V_w \to \mathbb{R}$,

$f: V_w \to \mathbb{R}$.

They are strictly increasing functions.

We can take for all components the underlimits $\Psi_T = \Psi_I = \Psi_F = -45$, and the overlimits $\Omega_T = \Omega_I = \Omega_F = 30$.

For the truth-value function, there exist the truth lower-threshold:

$\eta_T^L = 0$, such that $t(\eta_T^L) = t(0) = 0$,

and the truth upper-threshold

$\eta_T^U = 20$, such that $t(\eta_T^U) = t(20) = 1$.

In this example, the thresholds are the same for the indeterminate-value function, as indeterminate lower-threshold one has

$\eta_I^L = 0$, such that $i(\eta_I^L) = i(0) = 0$,

and as indeterminate upper-threshold one has

$\eta_I^U = 20$, such that $i(\eta_I^U) = i(20) = 1$.

And for the false-value function: there exists a false lower-threshold

$\eta_F^L = 0$, such that $f(\eta_F^L) = f(0) = 0$,

and a false upper-threshold

$\eta_F^U = 20$, such that $f(\eta_F^U) = f(20) = 1$.

Therefore, the three functions' formulas, after rescaling them, can be defined respectively as: for any $v \in V_W$, one gets

$t(v) = \frac{v}{20}$ (degree of membership);

$i(v) = \frac{v}{20}$ (degree of indeterminate-membership);

$f(v) = \frac{v}{20}$ (degree of nonmembership).

- Suppose Antoinette has produced exactly 25 electronic devices, 2 of her electronic devices are in pending (due to quality control; hence they are in indeterminate status), whence the neutrosophic overset value ($N_O$) of her is:

$N_O(\text{Antoinette}\langle 25, 2, 0\rangle) = \langle t(25), i(2), f(0)\rangle =$
$= \langle \frac{25}{20}, \frac{2}{20}, \frac{0}{20}\rangle = \langle 1.25, 0.10, 0\rangle,$
so she has an overmembership to the company *Inventica*. She is over-productive.

- Adriana, another employee, has produced 11 electronic devices, and one is in pending. Since the norm was 20, she missed $20 - 11 - 1 = 8$ electronic devices. Then

$N_o(\text{Adriana}\langle 11, 1, 8\rangle) = \langle \frac{11}{20}, \frac{1}{20}, \frac{8}{20}\rangle = \langle 0.55, 0.05, 0.40\rangle.$

So, her degree of membership is partial (0.55), her degree of indeterminate membership is 0.05, and her degree of nonmembership also partial (0.40).

- Oliver has tried to build 16 electronic devices, but he wrecked 10 of them, other 5 were successful, and from the left one he did only half. Another of his electronic devices.

Calculate: $10 \cdot (-2) = -20$ points. $(5 + 0.5) \cdot 1 = 5.5$ points.

$N_o(\text{Olivier}\langle -20, +5.5\rangle, 1, 3.5) = \langle t(-20) +$
$t(5.5), i(1), f(3.5)\rangle = \langle \frac{-20}{20} + \frac{5.5}{20}, \frac{1}{20}, \frac{3.5}{20}\rangle =$
$\langle -0.725, 0.050, 0.175\rangle$, so his degree of membership (contribution) to the company is negative.

- But Murriah has damaged 14 electronic devices, and 6 are still in the pending/indeterminate status since their quality is unclear.

Compute $14 \cdot (-2) = -28$ points. Then,

$N_o(\text{Murriah}\langle t(-28), i(6), f()0\rangle) = \langle \frac{-28}{20}, \frac{6}{20}, \frac{0}{20}\rangle =$
$\langle -1.4, 0.3, 0\rangle,$
so her membership degree of appurtenance to the company is negative, the worst so far! So, she is under-productive.

# World Companies as Neutrosophic Offsets

Actually most companies, institutions and associations have a structure of neutrosophic offsets, because they employ individuals:

- that work full-time (degree of membership = 1)
- that work part-time {degree of membership in (0, 1)}
- that work overtime (degree of membership > 1)
- that produce more damage than benefit to the company (destroying materials and tools, law suits, extended periods of absence, etc.) {degree of membership < 0}.

Also, the majority (if not all) of companies, institutions, associations and in general any real system is changing in time, or space, or regarding its structure and its composition, so they are **dynamic systems**, or better **neutrosophic dynamic systems**, and actually **neutrosophic dynamic offsystems**.

Thus, the previous example of company "Inventica", with its employees, is actually a neutrosophic dynamic offsystem.

A Neutrosophic System defined in "Symbolic Neutrosophic Theory" (2015), pp. 28-29, is a system which has some indeterminacy with respect to its space S, or to its elements, or at least one of its elements $x_o\left(t_{x_o}, i_{x_o}, f_{x_o}\right)$ do not 100% belong to $S$, with $\left(t_{x_o}, i_{x_o}, f_{x_o}\right) \neq (1, 0, 0)$, or at least one of its relationships $\mathcal{R}_o(t, i, f) \in S$, between its own elements, or betwen the system and the environment, are only partial relationships [i.e. $(t, i, f)$ -neutrosophically], with $(t, i, f) \neq (1, 0, 0)$.

A **Neutrosophic Offsystem** is a neutrosophic system which has at least one neutrosophic offelement, or has a neutrosophic overelement and a neutrosophic underelement.

Similarly, a **Neutrosophic Oversystem** is a neutrosophic system that has at least one neutrosophic overelement. And a **Neutrosophic Undersystem** is a neutrosophic system that has at least one underelement.

A **Neutrosophic Element** $x$ belongs to a neutrosophic set $A$ with a neutrosophic degree of membership $x\langle t_A, i_A, f_A \rangle \in A$ , where all neutrosophic components $t_A, i_A, f_A \subseteq [0,1]$.

A **Neutrosophic Offelement** $y$ belongs to a neutrosophic offset $O$ with a neutrosophic offdegree of membership $y\langle t_O, i_O, f_O \rangle \in O$ , such that one of the neutrosophic components $t_O, i_O, f_O$ are partially or totally above 1, and another neutrosophic component is partially or totally below 0.

A neutrosophic component, which is partially or totally above 1, is called **Neutrosophic Overcomponent**. And a neutrosophic component which is partially or totally below 0, is called **Neutrosophic Undercomponent**.

It is also possible to have a neutrosophic component which is both partially or totally above 1 and below 0, and it is called **Neutrosophic Offcomponent.** For example: the truth-value of the neutrosophic element $x \in U$, defined as: $t_x = [-0.1, 1.2]$.

# 28. *(t, i, f)*-Neutrosophic Offstructure

A $(\boldsymbol{t}, \boldsymbol{i}, \boldsymbol{f})$ −**Neutrosophic Offstructure** is a structure defined on a neutrosophic offset.

Similarly, a **(t, i, f)-Neutrosophic Overstructure** is a structure defined on a neutrosophic overset.

And a **(t, i, f)-Neutrosophic Understructure** is a structure defined on a neutrosophic underset.

We first recall the definition of a $(t, i, f)$ −*Neutrosophic Structure*[3]:

> Any structure is composed from two parts: a **space**, and a **set of axioms** (or **laws**) acting (governing) on it. If the space, or at least one of its axioms (laws) has some indeterminacy of the form $(t, i, f) \neq (1, 0, 0)$, that structure is a $(\boldsymbol{t}, \boldsymbol{i}, \boldsymbol{f})$ −**Neutrosophic Structure**.

Now, if there exist some indeterminacies of the form $(t_o, i_o, f_o)$ such that some neutrosophic components are partially or totally off the interval $[0, 1]$, both over and under [0, 1], then one has a $(\boldsymbol{t}, \boldsymbol{i}, \boldsymbol{f})$ − **Neutrosophic Offstructure**.

## Example 1 of *(t, i, f)*-Neutrosophic Overstructure

$\left(\mathbb{Z}_{(t,i,f)}^{(4)}, \boxplus\right)$ be the set generated by the element

$1_{(1.2,\,0.1,\,0.3)}$ modulo 4, with respect to the neutrosophic law

$$\boxplus: \mathbb{Z}_{(t,i,f)}^{(4)} \times \mathbb{Z}_{(t,i,f)}^{(4)} \to \mathbb{Z}_{(t,i,f)}^{(4)}$$

[3] Florentin Smarandache: *Symbolic Neutrosophic Theory*. EuropaNova: Brussels, Belgium, 2015; §4.1, p. 103.

$$x_{1(t_1,i_1,f_1)} \boxplus x_{2(t_2,i_2,f_2)}$$
$$= (x_1 + x_2)_{(max\{t_1,t_2\},min\{i_1,i_2\},min\{f_1,f_2\})}. \qquad (94)$$

Then:

$$1_{(1.2,0.1,0.3)} \boxplus 1_{(1.2,0.1,0.3)}$$
$$= (1 + 1)_{(max\{1.2,1.2\},min\{0.1,0.1\},min\{0.3,0.3\})}$$
$$= 2_{(1.2,0.1,0.3)}$$
$$2_{(1.2,0.1,0.3)} \boxplus 1_{(1.2,0.1,0.3)} = 3_{(1.2,0.1,0.3)}$$
$$3_{(1.2,0.1,0.3)} \boxplus 1_{(1.2,0.1,0.3)} = 4_{(1.2,0.1,0.3)}$$
$$\equiv 0_{(1.2,0.1,0.3)} (\text{mod } 4)$$

Hence

$$\mathbb{Z}_{(t,i,f)}^{(4)} = \{0_{(1.2,0.1,0.3)}, 1_{(1.2,0.1,0.3)}, 2_{(1.2,0.1,0.3)}, 3_{(1.2,0.1,0.3)},\}.$$

## Example 2

$\mathbb{Z}_{(t,i,f)}^{(3)}$ = the set generated by the elements $0_{(-0.1,0.1,0.7)}$ and $2_{(0.8,0.2,0.4)}$ modulo 3, with respect to the neutrosophic law:

$$\boxdot: \mathbb{Z}_{(t,i,f)}^{(3)} \times \mathbb{Z}_{(t,i,f)}^{(3)} \rightarrow \mathbb{Z}_{(t,i,f)}^{(3)},$$

defined as:

$$x_{1(t_1,i_1,f_1)} \boxdot x_{2(t_2,i_2,f_2)}$$
$$= (x_1 \cdot x_2)_{(min\{t_1,t_2\},max\{i_1,i_2\},max\{f_1,f_2\})} \qquad (95)$$

Then:

$$2_{(0.8,0.2,0.4)} \boxdot 2_{(0.8,0.2,0.4)}$$
$$= (2 \cdot 2)_{(min\{0.8,0.8\},max\{0.2,0.2\},max\{0.4,0.4\})}$$
$$= 4_{(0.8,0.2,0.4)} \equiv 1_{(0.8,0.2,0.4)}(mod\ 3)$$
$$0_{(-0.1,0.1,0.7)} \boxdot 2_{(0.8,0.2,0.4)}$$
$$= (0 \cdot 2)_{(min\{-0.1,0.8\},max\{0.1,0.2\},max\{0.7,0.4\})} = 0_{(-0.1,0.2,0.7)}$$
$$0_{(-0.1,0.1,0.7)} \boxdot 1_{(0.8,0.2,0.4)}$$
$$= (0 \cdot 1)_{(min\{-0.1,0.8\},max\{0.1,0.2\},max\{0.7,0.4\})}$$
$$= 0_{(-0.1,0.2,0.7)}$$

Since the neutrosophic membership degree of the element " 0 " is hesitating between $(-0.1, 0.1, 0.7)$ and $(-0.1, 0.2, 0.7)$, we conclude that

$$0_{(-0.1,\{0.1,0.2\},0.7)} \in \mathbb{Z}^{(3)}_{(t,i,f)}$$

Hence

$$\left(\mathbb{Z}^{(4)}_{(t,i,f)}, \boxplus\right) = \left\{0_{(-0.1,\{0.1,0.2\},0.7)}, 1_{(0.8,0.2,0.4)}, 2_{(0.8,0.2,0.4)}\right\}.$$

# 29. Neutrosophic Offprobability

The company manager of "Inventica" fires Murriah, because of her bad work and hires a new employee: Costel. What is the probability that Costel will be a good worker? If one says that $P$(Costel good worker) ∈ [0,1] as in classical probability, or $P$ (Costel good worker) ⊆ [0,1] as in classical imprecise probability, one obtains incomplete response, because the extremes exceeding 1 or below 0 are omitted. Costel can be an excellent worker, doing overload and producing above the required norm of 20 electronic devices per week, hence the neutrosophic offprobability

$NP_O$(Costel good worker) > 1,

or Costel can cause problems for the company by damaging electronic devices and tools, by law suits against the company etc., hence

$NP_O$(Costel good worker) < 0.

Therefore, we extend the classical probabilistic interval [0, 1] to the left and to the right sides, to

$$\left[\dots\frac{-15}{20},\frac{30}{20}\dots\right] = [\dots - 2.25, 1.50 \dots],$$

where the three dots "…" in each side mean that the underlimit and respectively overlimit of the interval are flexible (they may change in time).

The complete response is now:

$NP_O$(Costel good worker) ∈[… -2.25, 1.50 …]$^3$

if one uses crsip numbers, or:

$NP_O$(Costel good worker) ⊆[… -2.25, 1.50 …]$^3$

If one uses hesitant/interval-valued/subset-value neutrosophic offprobability.

# 30. Definition of Neutrosophic Offprobability

Let $\mathcal{S}$ be a Neutrosophic Probability Space (i.e., a probability space that has some indeterminacy).

The Neutrosophic Probability of an event $E \in \mathcal{S}$ is:

$$\left\langle \begin{matrix} \textit{the chance that} \\ E \textit{ occurs} \end{matrix} \quad \begin{matrix} \textit{the indeterminate chance} \\ \textit{about } E \textit{ occurence} \end{matrix} \quad \begin{matrix} \textit{the chance that} \\ E \textit{ does not occur} \end{matrix} \right\rangle = < ch(E), ch(neutE), ch(antiE) >. \qquad (96)$$

If there exist some events $E_1, E_2 \in \mathcal{S}$ such that two of their neutrosophic components $ch(E_1)$, or $ch(neutE_1)$, or $ch(antiE_1)$, or also $ch(E_2)$, or $ch(neutE_2)$, or $ch(antiE_2)$, are both partially or totally off the interval [0, 1], i.e. one of them above 1 and the other one below 0, one has a Neutrosophic Offprobability.

Similarly, a **Neutrosophic Overprobability** is a neutrosophic probability whose probability space has at least one event $E_0$ whose at least one neutrosophic component $ch(E_0)$, or $ch(neutE_o)$ , or $ch(antiE_0)$ is partially or totally above 1.

And a **Neutrosophic Underprobability** is a neutrosophic probability whose probability space has at least one event $E_0$ whose at least one neutrosophic component $ch(E_0)$, or $ch(neutE_o)$, or $ch(antiE_0)$ is partially or totally below 0.

# 31. Definition of Neutrosophic Offstatistics

Neutrosophic Statistics means statistical analysis of population sample that has indeterminate (imprecise, ambiguous, vague, incomplete, unknown) data. For example, the population of sample size might not be exactly determinate because of some individuals that partially belong to the population or sample, and partially they do not belong, or individuals whose appurtenance is completely unknown. Also, there are population or sample individuals whose data could be indeterminate.

Neutrosophic Offstatistics adds to this the existence of individuals that have on overmembership (i.e. membership > 1) to the population or sample, and an undermembership (i.e. membership < 0) to the population or sample.

Neutrosophic Offstatistics is connected with the Neutrosophic Offprobability, and it is an extension of the Neutrosophic Statistics[4].

Hence, **Neutrosophic Offstatistics** means statistical analysis of population or sample that has indeterminate (imprecise, ambiguous, vague, incomplete, unknown) data, when the population or sample size cannot be exactly determinate because of some individuals that partially belong and partially do not belong to the population or sample, or individuals whose appurtenance is completely unknown, and there are individuals that have an overappurtenance (degree of appurtenance > 1) and a underappurtenance (degree of appurtenance $< 0$ ). Also, there are population or sample individuals whose data could be indeterminate.

[4] Florentin Smarandache, *Introduction to Neutrosophic Statistics*, Sitech Craiova, 123 pages, 2014, http://fs.gallup.unm.edu/NeutrosophicStatistics.pdf

It is possible to define the neutrosophic offstatistics in many ways, because there are various types of indeterminacies, and many styles of overappurtenance / underappurtenance, depending on the problem to solve.

**Neutrosophic Overstatistics** is connected with the Neutrosophic Overprobability, and it studies populations or samples that contain individuals with overmembership (but no individuals with undermembership).

**Neutrosophic Understatistics** is connected with the Neutrosophic Underprobability, and it studies populations or samples that contain individuals with undermembership (but no individuals with overmembership).

## Example of Neutrosophic Offstatistics

The neutrosophic population formed by employees of company *Inventica*, from the previous example. Some employees have negative-appurtenance (contribution) to the company, others over-appurtenance, or partial-appurtenance i.e. in between [0, 1]. So, we deal with neutrosophic overstatistics. Let's take the following neutrosophic sample:

$A_{\text{NS}} = \{Antoinette, Adriana, Oliver, Murriah\}$.

We estimate the average of the whole population by the average of this sample.

The Neutrosophic averages of the sample is:

$$
\begin{aligned}
&\frac{1}{4}\cdot\big(\langle 1.25,0.10,0\rangle + \langle 0.55,0.05,0.40\rangle \\
&\qquad + \langle -0.725,0.050,0.175\rangle + \langle -1.4,0.3,0\rangle\big) \\
&= \frac{1}{4}\cdot\left\langle \begin{matrix} 1.25 + 0.55 + (-0.725) + (-1.4), \\ 0.10 + 0.05 + 0.3, \\ 0 + 0.40 + 0.175 + 0 \end{matrix} \right\rangle \\
&= \frac{1}{4}\cdot\langle -0.325,0.500,0.575\rangle \\
&= \left\langle \frac{-0.325}{4},\frac{0.500}{4},\frac{0.575}{4}\right\rangle \\
&= \langle -0.08125, 0.12500, 0.14375\rangle,
\end{aligned}
$$

which shows a negative contribution to the company. Therefore, many employees have to be let go, and devoted and carefully selected new employees should be hired.

# 32. Definition of Refined Neutrosophic Probability

Let $\mathcal{S}$ be a Neutrosophic Probability Space. Then, the refined neutrosophic probability that an event $E \in \mathcal{S}$ to occur is:

$$NP_R(E) = \begin{pmatrix} \langle ch_1(E)\rangle, \langle ch_2(E)\rangle, \dots, \langle ch_p(E)\rangle, \\ \langle ch_1(\text{neut}E)\rangle, \langle ch_2(\text{neut}E)\rangle, \dots, \langle ch_r(\text{neut}E)\rangle, \\ \langle ch_1(\text{anti}E)\rangle, \langle ch_2(\text{anti}E)\rangle, \dots, \langle ch_s(\text{anti}E)\rangle \end{pmatrix}, \quad (97)$$

$ch_j(E) =$ the subchance (or subprobability) of type $j$ that the event $E$ occurs, where $j \in \{1, 2, \dots, p\}$;

$ch_k(\text{neut}E) =$ the indeterminate-subchance (or indeterminate-subprobability) of type $k$ that the event $E$ occurs, where $k \in \{1, 2, \dots, r\}$;

$ch_l(\text{anti}E) =$ the subchance (or subprobability) of type $l$ that the event $E$ does not occur (or that the opposite of the event $E$, i.e. anti$E$, occurs), where $l \in \{1, 2, \dots, s\}$, with $p + r + s \geq 4$, all $ch_j(E)$, $ch_k(\text{neut}E)$, $ch_l(\text{anti}E) \subseteq [0, 1]$ for all $j$, $k$, and $l$.

Of course, the neutrosophic probability refinement can be done in many ways, for the same event, depending on the problem to solve and on the available data.

## Example of Refined Neutrosophic Probability

Suppose the event $E$ = "John candidates for the US Presidency in the next voting process".

$NP_R(E) = (\langle 0.2, 0.3\rangle, \langle 0.0, 0.1\rangle, \langle 0.3, 0.1\rangle)$,

where:

$$\begin{cases} ch_1(E) = 0.2 \\ ch_2(E) = 0.3 \end{cases}$$

$ch_1(E)$ represents the percentage of men from whole country that are likely to vote for John;

$ch_2(E)$ represents the percentage of women from whole country that are likely to vote for John.

$$\begin{cases} ch_1(\text{neut}E) = 0.2 \\ ch_2(\text{neut}E) = 0.3 \end{cases}$$

$ch_1(\text{neut}E)$ represents the percentage of men from whole country that are likely not to vote;

$ch_2(\text{neut}E)$ represents the percentage of women from whole country that are likely not to vote.

$$\begin{cases} ch_1(\text{anti}E) = 0.2 \\ ch_2(\text{anti}E) = 0.3 \end{cases}$$

$ch_1(\text{anti}E)$ represents the percentage of men from whole country that are likely to vote against John;

$ch_2(\text{anti}E)$ represents the percentage of women from whole country that are likely to vote against John.

# 33. Definition of Refined Neutrosophic Offprobability

It is defined similarly as the previous refined neutrosophic probability, with the condition that there exist some events $E_1, E_2 \in \mathcal{S}$ such that at least two of their neutrosophic subchances (subprobabilities):

$ch_1(E_j), ch_2(E_j), \dots, ch_p(E_j),$
$ch_1(neutE_j), ch_2(neutE_j), \dots, ch_r(neutE_j),$
$ch_1(antiE_j), ch_2(antiE_j), \dots, ch_s(antiE_j),$
for $j \in \{1, 2\}$, (98)

are partially or totally off the interval $[0, 1]$, i.e. one neutrosophic subchance above 1 and another neutrosophic subchance below 0.

Similarly, a **Refined Neutrosophic Over-probability** is a refined neutrosophic probability, such that at least one of its event has at least a neutrosophic subchance that is partially or totally above 1 (and there is no neutrosophic subchance partially or totally below 0).

Similarly, a **Refined Neutrosophic Under-probability** is a refined neutrosophic probability, such that at least one of its event has at least a neutrosophic subchance that is partially or totally below 0 (and there is no neutrosophic subchance partially or totally above 1).

# 34. Definition of Neutrosophic Offlogic

In the **Neutrosophic Propositional Logic**, to each proposition $\mathcal{P}$ one associates a triple $(T_p, I_p, F_p)$, and one says that the neutrosophic truth-value of the proposition $\mathcal{P}(T_p, I_p, F_p)$ is $T_p$ true, $I_p$ indeterminate, and $F_p$ false, where $T_p, I_p, F_p \subseteq [0, 1]$.

A neutrosophic proposition $\mathcal{P}_o\,(T_{p_o}, I_{p_o}, F_{p_o})$ is called **Neutrosophic Offproposition** if one neutrosophic component among $T_{p_o}, I_{p_o}, F_{p_o}$ is partially or totally above 1, and another one is partially or totally below 0. Or it has a neutrosophic offcomponent (i.e. a neutrosophiic component that is simultaneously above 1 and below 0, for example one of the form [-0.2, +1.1]).

A neutrosophic proposition $\mathcal{P}_o\,(T_{p_o}, I_{p_o}, F_{p_o})$ is called **Neutrosophic Overproposition** if one neutrosophic component among $T_{p_o}, I_{p_o}, F_{p_o}$ is partially or totally above 1, and there is no neutrosophic component that is partially or totally below 0.

A proposition $\mathcal{P}_o\,(T_{p_o}, I_{p_o}, F_{p_o})$ is called **Neutrosophic Underproposition** if one neutrosophic component among $T_{p_o}, I_{p_o}, F_{p_o}$ is partially or totally below 0, and there is no neutrosophic component that is partially or totally above 1.

A **Neutrosophic Offlogic** is a neutrosophic logic that has at least a neutrosophic offproposition.

A **Neutrosophic Overlogic** is a neutrosophic logic that has at least a neutrosophic overproposition, and has no neutrosophic underproposition.

A **Neutrosophic Underlogic** is a neutrosophic logic that has at least a neutrosophic underproposition, and has no neutrosophic overproposition.

## Example of Neutrosophic Offlogic

We return to the example with company *Inventica*. Let's identify an employee, Bruce, of this company.

Consider the proposition:

$Q$ = {Bruce does a good work for the company *Inventica*}.

What is the truth-value of this proposition?

To say that the truth-value of $Q$ belongs to [0, 1] in the case when one has crisp truth-value, or the truth-value of Q is included in or equal to [0, 1] when one has hesitant / interval / subset-value is, analogously to the case of neutrosophic offprobability, incomplete, because one misses the situation when Bruce does damage to the company:

$t(NL_o(Q)) < 0$, where $t(NL_o(Q))$ means the truth-value neutrosophic component,
and the case when Bruce does overload, hence

$t(NL_o(Q)) > 1$.

The complete response is:

$NL_o(Q) \in [\ldots -2.25, 1.50 \ldots]^3$ if one uses crisp neutrosophic offlogic, or

$NL_o(Q) \subseteq [\ldots -2.25, 1.50 \ldots]^3$ if one uses hesitant / interval / subset-value neutrosophic offlogic.

# 35. The Neutrosophic Offquantifiers

The **Neutrosophic Quantifiers**[5] are straightforwardly extended to neutrosophic logic in the following way:

1) **The Neutrosophic Existential Offquantifier**:

$$\exists x < t_x, i_x, f_x > \in A, P(x) < t_{P(x)}, i_{P(x)}, f_{P(x)} >, \qquad (99)$$

which means: there exists a neutrosophic element x that belongs to the neutrosophic overset A in the neutrosophic degree $< t_x, i_x, f_x >$, such that the proposition P(x) has the neutrosophic degree of truth $< t_{P(x)}, i_{P(x)}, f_{P(x)} >$, and at least one of the neutrosophic components $t_x, i_x, f_x$, $t_{P(x)}, i_{P(x)}, f_{P(x)}$ is partially or totally off the interval [0, 1].

2) **The Neutrosophic Universal Offquantifier**:

$$\forall x < t_x, i_x, f_x > \in A, P(x) < t_{P(x)}, i_{P(x)}, f_{P(x)} >, \qquad (100)$$

which means: for any neutrosophic element x that belongs to the neutrosophic overset A in the neutrosophic degree $< t_x, i_x, f_x >$, such that the proposition P(x) has the neutrosophic degree of truth $< t_{P(x)}, i_{P(x)}, f_{P(x)} >$, and at least one of the neutrosophic components $t_x, i_x, f_x$, $t_{P(x)}, i_{P(x)}, f_{P(x)}$ is partially or totally over 1, and another neutrosophic component of P(x) or of another proposition is partially or totally below 0.

---

[5] Florentin Smarandache: *Symbolic Neutrosophic Theory*. EuropaNova: Brussels, Belgium, 2015; §3.10, p. 71-72.

# 36. Definition of Refined Neutrosophic Offet

We introduce for the first time the Refined Neutrosophic Overset.

Let $\mathcal{U}$ be a universe of disourse, and let $O_R$ be a refined neutrosophic set of $\mathcal{U}$, i.e.

$O_R \subset \mathcal{U}$,

$$O_R = \left\{ \begin{matrix} x\left(T_{O_R}^{j}, I_{O_R}^{k}, F_{O_R}^{l}\right), \\ j \in \{1, 2, \dots, p\}, \\ k \in \{1, 2, \dots, r\}, \\ l \in \{1, 2, \dots, s\}, \\ p + r + s \geq 4, \\ x \in \mathcal{U} \end{matrix} \right\}, \tag{101}$$

where

$T_{O_R}^{j}$ is type $j$ of subtruth-submembership,

$I_{O_R}^{k}$ is type $k$ of subindeterminacy-submembership,

$F_{O_R}^{l}$ is type $l$ of subfalsehood-submembership,

of the generic element $x$ with respect to the set $O_R$.

We say that $O_R$ is a Refined Neutrosophic Overset if there exists at least one element

$$y\begin{pmatrix} T_y^{j}, I_y^{k}, F_y^{l};\ j \in \{1, 2, \dots, p\}, k \in \{1, 2, \dots, r\}, \\ l \in \{1, 2, \dots, s\}, p + r + s \geq 4 \end{pmatrix} \tag{102}$$

whose at least one subcomponent among all of them

$$\left(T_y^1, T_y^2, \dots, T_y^p;\ I_y^1, I_y^2, \dots, I_y^r;\ F_y^1, F_y^2, \dots, F_y^s\right) \tag{103}$$

is partially or totally over 1 and another component of y or of another element that is partially or totally below 0.

For example:

$$O_{\mathcal{R}} = \begin{Bmatrix} x_1(-0.1, 0.2; 0.3; 0.6, 0.5, 0.3), \\ x_2(0, 0.9; 0.2; 0.4, 1.1, 0.7) \end{Bmatrix},$$

where the first element has a negative degree of membership of type 1 (i.e. $T^1 = -0.1$), and the second element has an over 1 degree of nonmembership of type 2 (i.e. $F^2 = 1.1$).

# 37. Definition of Refined Neutrosophic Logic

Any logical proposition $Q$ has the degree $T_Q^j$ of subtruth of type $j$, for $j \in \{1, 2, \dots, p\}$, the degree $I_Q^k$ of subindeterminacy of type $k$ for $k \in \{1, 2, \dots, r\}$, and the degree $F_Q^l$ of subfalsehood of type $l$, for $l \in \{1, 2, \dots, l\}$,
with $p + r + s \geq 4$, and all $T_Q^j, I_Q^k, F_Q^l \subseteq [0, 1]$.

# 38. Definition of Refined Neutrosophic Overlogic

A refined neutrosophic logic, defined as above, with the condition that there exists at least one proposition

$$Q_O\left(T_{Q_O}^{j}, I_{Q_O}^{k}, F_{Q_O}^{l}\right),$$

such that at least one of its subcomponents

$$T_{Q_O}^{1}, T_{Q_O}^{2}, \dots, T_{Q_O}^{p}, I_{Q_O}^{1}, I_{Q_O}^{2}, \dots, I_{Q_O}^{r}, F_{Q_O}^{1}, F_{Q_O}^{2}, \dots, F_{Q_O}^{s}$$

is partially or totally over 1.

# 39. Definition of Refined Neutrosophic Underlogic

A refined neutrosophic logic, defined as above, with the condition that there exists at least one proposition

$$Q_O\left(T_{Q_O}^j, I_{Q_O}^k, F_{Q_O}^l\right),$$

such that at least one of its subcomponents

$$T_{Q_O}^1, T_{Q_O}^2, \dots, T_{Q_O}^p, I_{Q_O}^1, I_{Q_O}^2, \dots, I_{Q_O}^r, F_{Q_O}^1, F_{Q_O}^2, \dots, F_{Q_O}^s$$

is partially or totally below 0.

## 40. Definition of Refined Neutrosophic Offlogic

A refined neutrosophic logic, defined as above, which both neutrosophic overlogic and neutrosophic underlogic.

# 41. Definition of Refined Fuzzy Set

Let $\mathcal{U}$ be a universe of discourse, and let $A \subset \mathcal{U}$ be a fuzzy set, such that:

$$A = \{x(T_x^1, T_x^2, \ldots, T_x^p), p \geq 2, x \in \mathcal{U}\}, \quad (104)$$

where $T_x^1$ is a degree of subtruth-submembership of type 1 of element $x$ with respect to the fuzzy set $A$, $T_x^2$ is a degree of subtruth-submembership of type 2 of element $x$ with respect to the fuzzy set $A$, and so on, $T_x^p$ is a degree of subtruth-submembership of type $p$ of element $x$ with respect to the fuzzy set $A$, where all $T_x^j \subseteq [0, 1]$.

## Example of Refined Fuzzy Set

$$A = \{d(0.1, 0.2, 0.5), e(0.6, [0.1,0.2], \{0.6, 0.7\})\}.$$

# 42. Definition of Refined Fuzzy Offset

A refined fuzzy set $A_O$, as defined above, but with the condition that there are some elements that have at least one subcomponent, which is partially or totally over 1, and another subcomponent which is partially or totally below 0.

## Example of Refined Fuzzy Offset

$$B = \{u(-0.41, 0, 0.6, 0.2), v(0.7, 0.2, [0.9, 1.2], -0.11)\}.$$

# 43. Definition of Refined Fuzzy Logic

Any logical proposition Q has the degree $T_Q^1$ of subtruth of type 1, the degree $T_Q^2$ of subtruth of type 2, and so on, the degree $T_Q^p$ of subtruth of type $p$, where all $T_Q^j \subseteq [0, 1]$.

## 44. Definition of Refined Fuzzy OffLogic

A refined fuzzy logic as above, with the condition that there exist some logical propositions such that at least one of their subtruths is partially ortotally above 1, and another subtruth is partially or totally below 0.

# 45. Definition of Refined Intuitionistic Fuzzy Set

Let $\mathcal{U}$ be a universe of discourse, and let $C \subset \mathcal{U}$ be an intuitionistic fuzzy set, such that:

$$C = \{x(T_x^j, F_x^l)\}, \tag{105}$$

$j \in \{1, 2, \dots, p\}, l \in \{1, 2, \dots, s\}, p + s \geq 3, x \in \mathcal{U}$,

where $T_x^j$ is the type $j$ of subtruth-submembership of element $x$ with respect to the set $C$, and $F_x^l$ is the type $l$ of subfalsehood-subnonmembership of element $x$ with respect to the set $C$, with all $T_x^j, F_x^l \subseteq [0, 1]$, and

$$\sum_{j=1}^{p} supT_x^j + \sum_{l=1}^{s} supF_x^l \leq 1. \tag{106}$$

## Example of Refined Intuitionistic Fuzzy Set

$$C = \left\{ \begin{matrix} x(\langle 0.2, 0.3\rangle, \langle 0.1, 0.3, 0.0\rangle), \\ y(\langle 0.0, 0.4\rangle, \langle [0.1, 0.2], 0.3, 0.1\rangle) \end{matrix} \right\}.$$

# 46. Definition of Refined Intuitionistic Fuzzy Offset

A refined intuitionistic fuzzy set $C_O$, defined as above, with the condition that there exist some elements such that at least one subcomponent is partially or totally above 1, and another subcomponent is partially or totally below 0.

## Example of Refined Intuitionistic Fuzzy Offset

$$C_O = \begin{Bmatrix} z(\langle -0.7, 0.1, [0.2, 0.3] \rangle, \langle 0.6, 0.0 \rangle), \\ w(\langle 0.2, 0.3, 0.0 \rangle, \langle 0.1, 1.1 \rangle) \end{Bmatrix}.$$

# 47. Definition of Refined Intuitionistic Fuzzy Logic

Any logical proposition $Q$ has the degree $T_Q^j$ of subtruth of type $j$, for $j \in \{1, 2, \ldots, p\}$, and the degree $F_Q^l$ of subfalsehood of type $l$, for $l \in \{1, 2, \ldots, s\}$, with all $T_Q^j, F_Q^l \subseteq [0, 1]$, and $\sum_{j=1}^{p} supT_Q^j + \sum_{l=1}^{s} supF_Q^l \leq 1.$ (107)

# 48. Definition of Refined Intuitionistic Fuzzy OffLogic

A refined intuitionistic fuzzy logic, defined as above, with the condition that there exist some logical propositions such that at least one of their subcomponents is partially or totally above 1, and another subcomponent that is partially or totally below 0.

# 49. Neutrosophic Offset Operators

Let's consider a universe of discourse $\mathcal{U}$, and $O(\mathcal{U})$ all neutrosophic off-sets defined on $\mathcal{U}$, whose elements have the form:

$x(T_O, I_O, F_O)$,

where $T_O, I_O, F_O$ are real standard or nonstandard subsets as follows:

$$T_O \subseteq [^{-}\Psi_T, \Omega_T^{+}]$$
$$I_O \subseteq [^{-}\Psi_I, \Omega_I^{+}]$$
$$F_O \subseteq [^{-}\Psi_F, \Omega_F^{+}] \quad (108)$$

where $\Psi_T, \Psi_I, \Psi_F$ representing the ***lower tresholds*** of $T_O, I_O, F_O$ respectively, and $\Omega_T, \Omega_I, \Omega_F$ representing the ***upper tresholds*** of $T_O, I_O, F_O$ respectively.

We extend the neutrosophic N-norm and N-conorm to the **Neutrosophic N-offnorm** and **Neutrosophic N-offconorm** respectively.

Since the non-standard subsets do not have applications in technical, engineering and other practical problems, we do not use non-standard analysis next, but only real standard subsets, i.e.

$$T_O \subseteq [\Psi_T, \Omega_T],$$
$$I_O \subseteq [\Psi_I, \Omega_I],$$
$$F_O \subseteq [\Psi_F, \Omega_F], \quad (109)$$

such that each of them includes the classical interval $[0, 1]$. Therefore, $\Psi_T, \Psi_I, \Psi_F \leq 0$ and $\Omega_T, \Omega_I, \Omega_F \geq 1$.

There are three types of neutrosophic off-set operators (depending on each practical application to solve):

a. The case when the thresholds $\Psi$ and $\Omega$ prevail over the classical 0 and 1 respectively.

b. The case when the classical 0 and 1 prevails over the thresholds Ψ and Ω respectively.
c. The mixed case, i.e. when either the lower threshold Ψ prevails over 0, by the upper threshold Ω does not prevail over 1. Or the opposite.

More objective looks the first case, that we'll present in this research. The last two cases are rather subjective.

# 50. The Neutrosophic Component N-offnorm [the class of neutrosophic offAND operators]

Let's denote by " $c$ " a neutrosophic component (i.e. $T_O$, or $I_O$, or $F_O$),

$c: M_O \to [\Psi, \Omega]$ where $\Psi$ is its lower threshold, while $\Omega$ is its upper threshold with respect to each component.

The neutrosophic component N-offnorm,

$$N_O^n: [\Psi, \Omega]^2 \to [\Psi, \Omega]. \quad (110)$$

For any elements $x, y$ and $z \in M_O$ one has the following axioms:

i) Overbounding Conditions:

$$N_O^n(c(x), \Psi) = \Psi, N_O^n(c(x), \Omega) = c(x). \quad (111)$$

ii) Commutativity:

$$N_O^n(c(x), c(y)) = N_O^n(c(y), c(x)). \quad (112)$$

iii) Monotonicity: If c(x) ≤ c(y), then

$$N_O^n(c(x), c(z)) \leq N_O^n(c(y), c(z)). \quad (113)$$

iv) Associativity:

$$N_O^n(N_O^n(c(x), c(y)), c(z)) = N_O^n(c(x), N_O^n(c(y), c(z))) \quad (114)$$

For simplicity, instead of $N_O^n(c(x), c(y))$ will be using $c(x) \underset{O}{\wedge} c(y)$.

We extend the most used neutrosophic AND operator

$<T_1, I_1, F_1> \wedge <T_2, I_2, F_2> = <T_1 \wedge T_2, I_1 \vee I_2, F_1 \vee F_2>$

to a neutrosophic offAND operator:

$$<T_1, I_1, F_1> \underset{O}{\wedge} <T_2, I_2, F_2> = <T_1 \underset{O}{\wedge} T_2, I_1 \underset{O}{\vee} I_2, F_1 \underset{O}{\vee} F_2> \quad (115)$$

# 51. The Neutrosophic Component N-offconorm

[the class of neutrosophic offOR operators]

The neutrosophic component N-offconorm,
$N_O^{co}: [\Psi, \Omega]^2 \rightarrow [\Psi, \Omega].$ (116)
For any elements $x, y$ and $z \in M_O$ one has the following axioms:

v) Overbounding Conditions:
$N_O^{co}(c(x), \Omega) = \Omega, N_O^{co}(c(x), \Psi) = c(x)$. (117)

vi) Commutativity:
$N_O^{co}(c(x), c(y)) = N_O^{co}(c(y), c(x))$. (118)

vii) Monotonicity: If c(x) ≤ c(y), then
$N_O^{co}(c(x), c(z)) \leq N_O^{co}(c(y), c(z))$. (119)

viii) Associativity:
$N_O^{co}(N_O^{co}(c(x), c(y)), c(z)) = N_O^{co}(c(x), N_O^{co}(c(y), c(z)))$
(120)

Again, for simplicity, instead of $N_O^{co}(c(x), c(y))$ will be using $c(x) \underset{O}{\vee} c(y)$.

We extend the most used neutrosophic OR operator
$<T_1, I_1, F_1> \vee <T_2, I_2, F_2> = <T_1 \vee T_2, I_1 \wedge I_2, F_1 \wedge F_2>$

to a neutrosophic offOR operator:

$<T_1, I_1, F_1> \underset{O}{\vee} <T_2, I_2, F_2> = <T_1 \underset{O}{\vee} T_2, I_1 \underset{O}{\wedge} I_2, F_1 \underset{O}{\wedge} F_2>$. (121)

**Remark.**

Among the well-known fuzzy set/logic T-norms / T-conorms, only the min / max respectively work for the neutrosophic offAND / offOR operators. Thus:

$c(x) \underset{O}{\wedge} c(y) = \min\{c(x), c(y)\}$ and

$$c(x) \underset{O}{\vee} c(y) = \max\{c(x), c(y)\}. \qquad (122)$$

The Algebraic Product T-norm / T-conorm

{i.e. T-norm(x, y) = x·y and T-conorm(x, y) = x + y - x·y} fail completely.

While the Bounded T-norm / T-conorm

{i.e. T-norm(x, y) = max{0, x + y - 1} and T-conorm(x, y) = min{1, x + y}}

can be upgraded to the neutrosophic offAND / offOR operators by substituting "0" with "Ψ", and "1" with "Ω". So, one gets:

$c(x) \underset{O}{\wedge} c(y) = \max\{\Psi, c(x) + c(y) - \Omega\}$ and

$$\text{and } c(x) \underset{O}{\vee} c(y) = \min\{\Omega, c(x) + c(y)\}. \qquad (123)$$

# 52. The Neutrosophic Offcomplement (Offnegation)

There is a class of such neutrosophic offcomplements. Therefore, the neutrosophic offcomplement of <T, I, F> can be:

either $<F, \Psi_I + \Omega_I - I, T>$

or $< \Psi_T + \Omega_T - T, I, \Psi_F + \Omega_F - F >$

or $< \Psi_T + \Omega_T - T, \Psi_I + \Omega_I - I, \Psi_F + \Omega_F - F>$ (124)

etc.

It is remarkable to know that the classical fuzzy complement:

$$\mathcal{C}(T) = 1 - T \quad (125)$$

where "T" is of course the truth value, is replaced in the neutrosophic offcomplement by:

$$C_O(T) = \Psi_T + \Omega_T - T \quad (126)$$

And similarly for the other two neutrosophic components:

$$C_O(I) = \Psi_I + \Omega_I - I, \quad (127)$$

$$C_O(F) = \Psi_F + \Omega_F - F. \quad (128)$$

This is done for the following raison:

$C_O(\Omega_T) = \Psi_T$ (the complement / opposite of the largest value is the smallest value);

$C_O(\Psi_T) = \Omega_T$ (the complement / opposite of the smallest value is the largest value);

and $C_O(a_T) = \Psi_T + \Omega_T - a_T$, for $a_T \in [\Psi_T, \Omega_T]$:

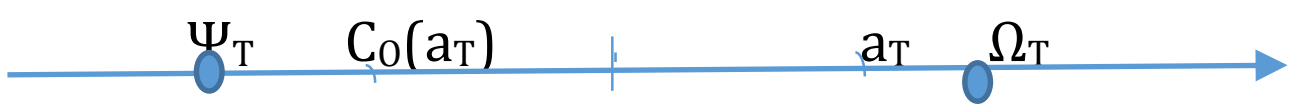

Fig. 2

$$\frac{\Psi_T + \Omega_T}{2}$$

In other words, the distance between "$a_T$" and the midpoint of the interval [$\Psi_T$, $\Omega_T$], which is $\frac{\Psi_T + \Omega_T}{2}$, is the same as the distance between $C_O(a_T)$ and that midpoint, i.e.

$$a_T - \frac{\Psi_T + \Omega_T}{2} = \frac{\Psi_T + \Omega_T}{2} - C_O(a_T) \quad (129)$$

or: $a_T + C_O(a_T) = \Psi_T + \Omega_T$. (130)

For $C_O(a_I)$ and $C_O(a_F)$ there are similar explanations.

In fuzzy set / logic, the property is the same:

$C(a) = 0 + 1 – a = 1 – a$, where $0 = \Psi_T$ and $1 = \Omega_T$,

and $a + C(a) = 0 + 1 = 1$,

also "a" and "C(a)" are at an equal distance from the midpoint of the interval [0, 1], which is 0.5.

An example: $C(0.7) = 1 – 0.7 = 0.3$,

but both numbers "0.7" and "0.3" are at the same distance from the midpoint 0.5.

## Example of Neutrosophic Offset Operators

Let's consider the single-valued neutrosophic components:

t, i, f: [-1.2, 1.2]

where, for all neutrosophic components, the lower threshold $\Psi$ = - 1.2, and the upper threshold $\Omega$ = + 1.2.

Let's suppose one has the following neutrosophic offsets:
A = {$x_1$<-1.1, 0.8, 0.9>, $x_2$<0.3, 0.6, 1.2>} and

B = {$x_1$<0.6, 1.1, -0.2>, $x_2$<0.3, 0.5, 0.7>}.

The neutrosophic offnegation of A is:

$\underset{O}{\neg} A = \{\underset{O}{\neg}[x_1<-1.1, 0.8, 0.9>], \underset{O}{\neg}[x_2<0.3, 0.6, 1.2>]\}$

$= \{\underset{O}{\neg} x_1<0.9, -1.2+1.2-0.8, -1.1>], \underset{O}{\neg}[x_2<1.2, -1.2+1.2-0.6,$

$0.3>]\} = \{\underset{O}{\neg} x_1<0.9, -0.8, -1.1>], \underset{O}{\neg}[x_2<1.2, -0.6, 0.3>]\}.$

i) The Neutrosophic Offintersection and Offunion:
Using the neutrosophic min / max offset operators:

$A \underset{O}{\cap} B = \{ x_1[<-1.1, 0.8, 0.9> \underset{O}{\wedge} <0.6, 1.1, -0.2>],$

$x_2[<0.3, 0.6, 1.2> \underset{O}{\wedge} <0.3, 0.5, 0.7>]\} = \{<\min\{-1.1,0.6\},$

$\max\{0.8,1.1\}, \max\{0.9, -0.2\}>, <\min\{0.3, 0.3\}, \max\{0.6,0.5\},$
$\max\{1.2, 0.7\}>\} = \{ x_1<-1.1, 1.1, 0.9>, x_2<0.3, 0.6, 1.2>\}.$

$A \underset{O}{\cup} B = \{ x_1[<-1.1, 0.8, 0.9> \underset{O}{\vee} <0.6, 1.1, -0.2>],$

$x_2[<0.3, 0.6, 1.2> \underset{O}{\vee} <0.3, 0.5, 0.7>]\} = \{x_1<\max\{-1.1, 0.6\},$

$\min\{0.8, 1.1\}, \min\{0.9, -0.2\}>, x_2<<\max\{0.3, 0.3\}, \min\{0.6, 0.5\},$
$\min\{1.2, 0.7\}>\} = \{x_1<0.6, 0.8, -0.2>, x_2<0.3, 0.5, 0.7>\}.$

ii) The Neutrosophic Offintersection and Offunion: Using the Bounded Neutrosophic N-offnorm / N-offconorm

In our example, one now has:

$$c(x) \underset{O}{\wedge} c(y) = \max\{-1.2, c(x) + c(y) - 1.2\},$$

and $$c(x) \underset{O}{\vee} c(y) = \min\{1.2, c(x) + c(y)\}.$$

$$A \underset{O}{\cap} B = \{ x_1[<-1.1, 0.8, 0.9> \underset{O}{\wedge} <0.6, 1.1, -0.2>],$$

$x_2[<0.3, 0.6, 1.2> \underset{O}{\wedge} <0.3, 0.5, 0.7>]\} = \{x_1<\max\{-1.2, -1.1+0.6-1.2\}, \min\{1.2, 0.8+1.1\}, \min\{1.2, 0.9+(-0.2)\}>, x_2<\max\{-1.2, 0.3+0.3-1.2\}, \min\{1.2, 0.6+0.5\}, \min\{1.2, 1.2+0.7\}>\} = \{x_1<-1.2, 1.2, 0.7>, x_2<-0.6, 1.1, 1.2>\}$.

$$A \underset{O}{\cup} B = \{ x_1[<-1.1, 0.8, 0.9> \underset{O}{\vee} <0.6, 1.1, -0.2>],$$

$x_2[<0.3, 0.6, 1.2> \underset{O}{\vee} <0.3, 0.5, 0.7>]\} = \{x_1<\min\{1.2, -1.1+0.6\}, \max\{-1.2, 0.8+1.1-1.2\}, \max\{-1.2, 0.9+(-0.2)-1.2\}>, x_2<\min\{1.2, 0.3+0.3\}, \max\{-1.2, 0.6+0.5-1.2\}, \max\{-1.2, 1.2+0.7-1.2\}>\} = \{x_1<-0.5, 0.7, -0.5>, x_2<0.6, -0.1, 0.7>\}$.

# 53. Application to Dynamic Systems

Most of the classical dynamic systems are actually neutrosophic dynamic systems on offsets, since besides elements that partially or totally belong to the system, there are elements with negative appurtenance (those that produce more damage than benefit to the system's functionality), as well as elements that are overloaded (i.e. those that produce more than the required full-time attribution norm).

# 54. Neutrosophic Tripolar (and Multipolar) Offset

We now introduce for the first time the neutrosophic tripolar overset, respectively the neutrosophic multipolar overset.

Let's start with an easy pratical example.

Suppose one has three universities, Alpha, Beta and Gamma, where a full-time student enrolles in 15 credit/hours and the maximum overload allowed is 18 credit hours.

University Alpha is competing 100% with University Beta in attracting students, since these universities offer the same courses and programs of studies. But University Gamma offers a totally different range of courses and programs of studies.

If John enrolls at the University Alpha in 6 credit hours, while other 3 credit hours are pending upon financial aid approval, then one has John's membership with respect to Alpha,

$John_{Alpha}\left(\frac{6}{15},\frac{3}{15},\frac{9}{15}\right)$.

But John enrolling in Alpha's studies is lost by the competing (opposite) University Beta, hence John's membership with respect to Beta is:

$John_{Beta}\left(-\frac{6}{15},-\frac{3}{15},-\frac{9}{15}\right)$,

while John's membership with respect to the University Gamma is not affected by him enrolling in Alpha or Beta, since the University Gamma is kind of neutral with respect to Alpha and Beta. Therefore one has:

$$John_{Gamma}\left(\frac{0}{15},\frac{0}{15},\frac{18}{15}\right).$$

Similarly, if another student, George, enrolls to the University Beta in credit units, while other 6 credit units, being pending (indeterminate), as:

$$George_{Beta}\left(\frac{9}{15},\frac{6}{15},\frac{3}{15}\right),$$

where

$$George_{Alpha}\left(-\frac{9}{15},-\frac{6}{15},-\frac{3}{15}\right),$$

and

$$George_{Gamma}\left(\frac{0}{15},\frac{0}{15},\frac{18}{15}\right).$$

The third student, Howard, enrolls to the University Gamma in 3 credit hours, while 9 credit hours being pending, or

$$Howard_{Gamma}\left(\frac{3}{15},\frac{9}{15},\frac{6}{15}\right),$$

where

$$Howard_{Alpha}\left(\frac{0}{15},\frac{0}{15},\frac{18}{15}\right),$$

and

$$Howard_{Beta}\left(\frac{0}{15},\frac{0}{15},\frac{18}{15}\right),$$

since universities Alpha and Beta are not affected by a student enrolled in Gamma.

We get the following table:

| University Alpha (+) | University Gamma (0) | University Beta (-) |
|---|---|---|
| $John_{Alpha}\left(\frac{6}{15},\frac{3}{15},\frac{9}{15}\right)$ | $John_{Gamma}\left(\frac{0}{15},\frac{0}{15},\frac{18}{15}\right)$ | $John_{Beta}\left(-\frac{6}{15},-\frac{3}{15},-\frac{9}{15}\right)$ |
| $George_{Alpha}\left(-\frac{9}{15},-\frac{6}{15},-\frac{3}{15}\right)$ | $George_{Gamma}\left(\frac{0}{15},\frac{0}{15},\frac{18}{15}\right)$ | $George_{Beta}\left(\frac{9}{15},\frac{6}{15},\frac{3}{15}\right)$ |
| $Howard_{Alpha}\left(\frac{0}{15},\frac{0}{15},\frac{18}{15}\right)$ | $Howard_{Gamma}\left(\frac{3}{15},\frac{9}{15},\frac{6}{15}\right)$ | $Howard_{Beta}\left(\frac{0}{15},\frac{0}{15},\frac{18}{15}\right)$ |

*Table 1*

Putting all three memberships tgether with respect to the three univesities <Alpha, Beta, Gamma >, where Alpha and Beta are 100% opposed to each other, while Gamma is completely neutral (100% independent) from ALpha and Beta, one has:

$$\text{John}\left(\langle\frac{6}{15},\frac{0}{15},-\frac{6}{15}\rangle,\langle\frac{3}{15},\frac{0}{15},-\frac{3}{15}\rangle,\langle\frac{9}{15},\frac{18}{15},-\frac{9}{15}\rangle\right),$$

$$\text{George}\left(\langle-\frac{9}{15},\frac{0}{15},\frac{9}{15}\rangle,\langle-\frac{6}{15},\frac{0}{15},\frac{6}{15}\rangle,\langle-\frac{3}{15},\frac{0}{15},\frac{3}{15}\rangle\right),$$

$$\text{Howard}\left(\langle\frac{0}{15},\frac{3}{15},\frac{0}{15}\rangle,\langle\frac{0}{15},\frac{9}{15},\frac{0}{15}\rangle,\langle\frac{18}{15},\frac{6}{15},\frac{18}{15}\rangle\right).$$

# 55. Degree of Anthagonism 100% Between Two Neutrosophic Offsets

We introduce for the first time the degree of anthagonism between two neutrosophic offsets.

Let $\mathcal{U}$ be a universe of discourse.

Two neutrosophic ofsets $O^+$ and $O^-$ are in degree of 100% anthagonism ($a^O = 1$) in the following case:

If $x(t_x, i_x, f_x) \in O^+$, then $x(-t_x, -i_x, -f_x) \in O^-$,     (131)

and reciprocally:

if $x(-t_x, -i_x, -f_x) \in O^-$, then $x(t_x, i_x, f_x) \in O^+$,     (132)

for any $x \in \mathcal{U}$ and $t_x, i_x, f_x \subseteq [\Psi, \Omega]$ .

For example, the above universities Alpha and Beta are in an anthagonism $a^0 = 1$.

# 56. General Definition of Neutrosophic Tripolar Offset

Let's consider three neutrosophic offsets $O^+, O^0$, and $O^-$, where $a^0(O^+, O^-) = 1$, meaning that the degree of anthagonism between $O^+$ and $O^-$ is 100%, and $a^0(O^+, O^0) = 0$, meaning that the degree of anthagonism between $O^+$ and $O^0$ is 0 (zero), and similarly the degree of anthagonism between $O^-$ and $O^0$ is 0 (zero).

Let's consider a universal set U. Then for the neutrosophic tripolar offset $O^+ \times\ O^0 \times\ O^-$ one has:
for each x ∈ U, x has the neutrosophic tripolar form:

$$x(<T_x^+, T_x^0, T_x^->, <I_x^+, I_x^0, I_x^->, <F_x^+, F_x^0, F_x^->)$$

where $x(<T_x^+, I_x^+, F_x^+>) \in O^+$,

$x(<T_x^0, I_x^0, F_x^0>) \in O^0$, and $x(<T_x^-, I_x^-, F_x^->) \in O^-$. (133)

See previous example with universities Alpha, Gamma, and respectively Beta.

# 57. General Degree of Anthagonism between Two Offsets

Let $\mathcal{U}$ be a universe of discourse.

We say that the **degree of anthagonism** between the neutrosophic offsets $O_a^+$ and $O_a^-$ is $a^O \in (0, 1)$ if:

for any $x \in \mathcal{U}$,

with $x(T_x^+, I_x^+, F_x^+) \in O_a^+$, and $x(T_x^-, I_x^-, F_x^-) \in O_a^-$,

one has:

$$\begin{cases} T_x^- = (-1) \cdot a^O \cdot T_x^+ \\ I_x^- = (-1) \cdot a^O \cdot I_x^+ \\ F_x^- = -[\Omega_F - a^O \cdot T_x^+ - a^O \cdot I_x^+] \\ \quad = -[\Omega_F - a^O(T_x^+ + I_x^+)] \\ \quad = -\Omega_F + a^O(T_x^+ + I_x^+) \end{cases} \tag{134}$$

**Example of Degree of Anthagonism**

$John_{Alpha}\left(\frac{6}{15}, \frac{3}{15}, \frac{9}{15}\right)$.

But the University Alpha is in a degree of anthagonism with University Delta, a fourth university, of

$a^O$(Alpha, Delta) = 0.8. Hence,

$$John_{Delta}\left(-1 \cdot (0.8) \cdot \frac{6}{15}, -1 \cdot (0.8) \cdot \frac{3}{15}, -\frac{18}{15} + 0.8\left(\frac{6}{15} + \frac{3}{15}\right)\right) =$$

$$John_{Delta} = \left(-\frac{4.8}{15}, -\frac{2.4}{15}, -\frac{10 \cdot 8}{15}\right).$$

# 58. Neutrosophic Multipolar Offset

In general, one has:

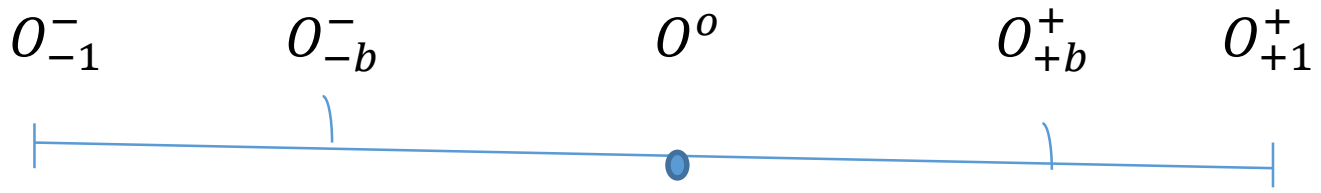

*Fig. 3*

where

$$a^o(O^-_{-1}, O^+_{+1}) = 1,$$

$$a^o(O^-_{-b}, O^+_{+b}) = 1,$$

$$a^o(O^-_{-1}, O^o) = a^o(O^-_{-b}, O^o) = a^o(O^+_{+b}, O^o) = a^o(O^+_{+1}, O^o) = 0, \quad (135)$$

and for any $b \in (0, 1)$, one has:

$$a^o(O^+_{+1}, O^-_{-b}) = a^o(O^-_{-1}, O^+_{+b}) = b \in (0, 1). \quad (136)$$

# 59. General Definition of Neutrosophic Multipolar Offset

Let's consider the neutrosophic offsets

$$O^+_{b_1}, O^+_{b_2}, \dots, O^+_{b_n}, O^0, O^-_{-b_n}, \dots, O^-_{-b_2}, O^-_{-b_1} \quad (137)$$

with $b_1, b_2, \dots, b_n \in (0,1), n \geq 1, b_1 < b_2 < \cdots < b_n$.

Let $\mathcal{U}$ be a universe of discourse.

One forms the neutrosophic multipolar offset:

$$O^+_{b_1} \times O^+_{b_2} \times \dots \times O^+_{b_n} \times O^0 \times O^-_{-b_n} \times \dots \times O^-_{-b_2} \times O^-_{-b_1} \quad (138)$$

and for each $x \in \mathcal{U}$, $x$ has the neutrosophic multipolar offset form:

$$x\begin{pmatrix} \langle T^+_1, T^+_2, \dots, T^+_n; T^0; T^-_{-n}, \dots, T^-_{-2}, T^-_{-1}\rangle, \\ \langle I^+_1, I^+_2, \dots, I^+_n; I^0; I^-_{-n}, \dots, I^-_{-2}, I^-_{-1}\rangle, \\ \langle F^+_1, F^+_2, \dots, F^+_n; F^0; F^-_{-n}, \dots, F^-_{-2}, F^-_{-1}\rangle \end{pmatrix}, \quad (139)$$

where $x < T^+_j, I^+_j, F^+_j > \in O^+_{+b_j}$,

and $x < T^-_{-j}, I^-_{-j}, F^-_{-j} > \in O^-_{-b_j}$, for j ∈{1, 2, ..., n},

while $x < T^0, I^0, F^0 > \in O^0$.

# 60. Particular Cases of Neutrosophic Multipolar Offset

1) The neutral $O^0$ may be removed from the above Cartesian product in certain applications, having only:

$$O_{b_1}^{+} \times O_{b_2}^{+} \times \ldots \times O_{b_n}^{+} \times O_{-b_n}^{-} \times \ldots \times O_{-b_2}^{-} \times O_{-b_1}^{-} \quad (140)$$

2) In the first Cartesian product one may not neccessarily need to have the same number of positive neutrosophic offsets $O_{b_j}^{+}$ as the number of negative neutrosophic offsets $O_{-b_k}^{-}$.

**Remark 1.**
One similarly can define, for the first time, the Fuzzy Tripolar Set / Offset and respectively Fuzzy Multipolar Set / Offset {just removing the neutrosophic components „I” (when I = 0) and „F”, and keeping only the first neutrosophic component „T”.

**Remark 2**.
Of course, one can also define, for the first time, the Intuitionistic Fuzzy Tripolar Set / Offset, and respectively the Intuitionistic Fuzzy Set / Offset by only removing the neutrosophic component „I” (when I = 0), and keeping the neutrosophic components „T” and „F”.

# 61. Symbolic Neutrosophic Offlogic

The **Symbolic Neutrosophic Offlogic Operators** (or we can call them Symbolic Neutrosophic Offoperators) are extensions of Symbolic Neutrosophic Logic Operators. The distinction is that for each symbolic neutrosophic component T, I, F, one has an over & under version:

$T_O$ = Over Truth,

$T_U$ = Under Truth;

$I_O$ = Over Indeterminacy,

$I_U$ = Under Indeterminacy;

$F_O$ = Over Falsehood,

$F_U$ = Under Falsehood.

# 62. Neutrosophic Symbolic Offnegation (Offcomplement)

| $\underset{O}{\neg}$ | $T_O$ | $T$ | $T_U$ | $I_O$ | $I$ | $I_U$ | $F_O$ | $F$ | $F_U$ |
|---|---|---|---|---|---|---|---|---|---|
| | $T_U$ | $F$ | $T_O$ | $I_U$ | $I$ | $I_O$ | $F_U$ | $T$ | $F_O$ |

*Table 2*

The neutrosophic offnegation of "over" component is the "under" component, and reciprocally.

$$\underset{O}{\neg}(T_O) = T_U \text{ and } \underset{O}{\neg}(T_U) = T_O. \quad (141)$$

$$\underset{O}{\neg}(I_O) = I_U \text{ and } \underset{O}{\neg}(I_U) = I_O. \quad (142)$$

$$\underset{O}{\neg}(F_O) = F_U \text{ and } \underset{O}{\neg}(F_U) = F_O. \quad (143)$$

The others remain the same as in symbolic neutrosophic logic:

$$\underset{O}{\neg}(T) = F,\ \underset{O}{\neg}(F) = T \text{ and } \underset{O}{\neg}(I) = I. \quad (144)$$

# 63. Symbolic Neutrosophic Offconjugation and Offdisjunction

For the Symbolic Neutrosophic Offconjugation and Symbolic Neutrosophic Offdisjunction, we need to define an order on the set of neutrosophic symbols

$$S_N = \{T_O, T, T_U, I_O, I, I_U, F_O, F, F_U\}. \quad (145)$$

The total or partial order defined on $S_N$ is not unique. It may depend on the application, or on the expert's believe, or if one uses the neutrosophic offlogic or neutrosophic offset or neutrosophic offprobability.

Let the relation of order ">" mean "more important than". We consider that T > I > F, hence T(ruth) is more important than I(ndeterminacy), which is more important than F(alsehood). Or F < I < T.

Then similarly: $T_O > I_O > F_O$ for the neutrosophic overcomponents that are bigger than 1, or $F_O < I_O < T_O$, whence one consequently deduces the neutrosophic undercomponents, which are < 0, if we multiply by -1 the previous double inequality; so, one gets: $T_U < I_U < F_U$.

Let's illustrate $S_N$ and its subjective order we defined, as follows:

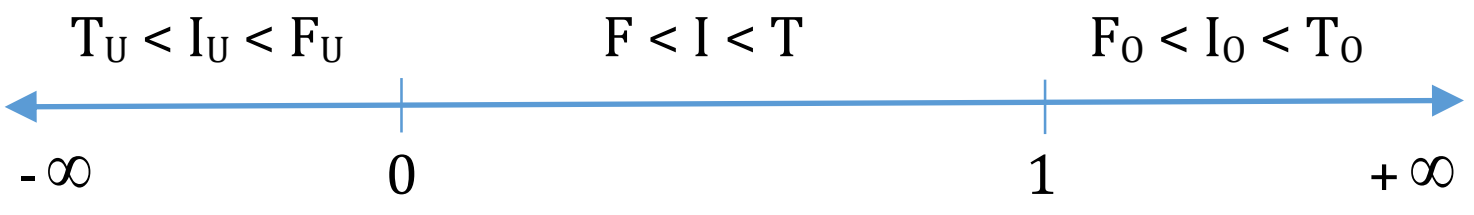

*Fig. 4*

which can be read in this way:

$T_U$, $I_U$, $F_U$ are under 0; F, I, T are between 0 and 1; while

$F_0$, $I_0$, $T_0$ are over 1.

$$T_U < I_U < F_U < F < I < T < F_0 < I_0 < T_0 \quad (146)$$

that is a total order on $S_N$.

Simply, one now defines the symbolic neutrosophic operators.

# 64. Symbolic Neutrosophic Offcomplement (Offnegation)

Remarkably, the symbolic neutrosophic offnegation (offcomplement) holds in the below order as in classical negation.

For each $\alpha \in S_N$ one has, the symbolic neutrosophic offcomplement $\mathcal{C}_O(\alpha)$ = the symmetric of α with respect to the median „I" in the symbolic sequence:

$$T_U, I_U, F_U, F, \mathbf{I}, T, F_O, I_O, T_O$$

We get the same results as above:

$\mathcal{C}_O(F_O) = F_U$, since $F_O$ and $F_U$ are symmetric with respect to „I".

$\mathcal{C}_O(F) = T$, for the same reason, etc.

# 65. Symbolic Neutrosophic Offconjunction (OffAND, or Offintersection)

For any $\alpha, \beta \in S_N$ one has

$$\alpha \underset{O}{\wedge} \beta = \min\{\alpha, \beta\} \qquad (147)$$

For examples:

$$T \underset{O}{\wedge} T_O = T \qquad (148)$$

$$I \underset{O}{\wedge} F = F \qquad (149)$$

$$F_U \underset{O}{\wedge} F_O = F_U \qquad (150)$$

$$I_U \underset{O}{\wedge} F = I_U \qquad (151)$$

$$T_U \underset{O}{\wedge} F_O = T_u \qquad (152)$$

# 66. Symbolic Neutrosophic Offdisjunction (OffOR, or Offunion)

For any $\alpha, \beta \in S_N$ one has

$$\alpha \underset{O}{\vee} \beta = \max\{\alpha, \beta\} \quad (153)$$

For examples:

$$T_U \underset{O}{\vee} F = F \quad (154)$$

$$I \underset{O}{\vee} I_0 = I_0 \quad (155)$$

$$T \underset{O}{\vee} F = T \quad (156)$$

$$F \underset{O}{\vee} T_0 = T_0 \quad (157)$$

# 67. Symbolic Neutrosophic Offimplication (Offinclusion)

For any $\alpha, \beta \in S_N$ one has:

$$\alpha \underset{O}{\rightarrow} \beta = \max\{\underset{O}{\neg}\alpha, \beta\}. \quad (158)$$

Examples:

$$I_O \underset{O}{\rightarrow} F = \max\{\underset{O}{\neg} I_O, F\} = \max\{I_U, F\} = F. \quad (159)$$

$$T \underset{O}{\rightarrow} T_O = \max\{\underset{O}{\neg} T, T_O\} = \max\{F, T_O\} = T_O. \quad (160)$$

$$F_U \underset{O}{\rightarrow} F_O = \max\{\underset{O}{\neg} F_U, F_O\} = \max\{F_O, F_O\} = F_O. \quad (161)$$

# 68. Symbolic Neutrosophic Offequivalence (Offequality)

Let $P$ and $Q$ be two offpropositions constructed with the neutrosophic symbols from the set $S_N$, together with the neutrosophic offoperators defined previously:

$$\neg_O, \wedge_O, \vee_O, \rightarrow_O. \quad (162)$$

Then we say that " $P \leftrightarrow_O Q$ " for the symbolic neutrosophic offlogic if $P \rightarrow_O Q$ and $Q \rightarrow_O P$.

Similarly, for the symbolic neutrosophic offset, let $P$ and $Q$ be offsets formed by the symbols of $S_N$ and with previously defined neutrosophic operators: $\mathcal{C}_O$ (complement), $\cap_O$, $\cup_O$ and $\subset_O$.

Then, we say that $P = Q$ for the symbolic neutrosophic offsets, if $P \subseteq_O Q$ and $Q \subseteq_O P$.

# 69. Different Symbolic Total Order

The readers may come up with a different symbolic total order on $S_N$. For example, starting from $T > F > I$, and doing a similar extension, one gets another neutrosophic total order on $S_N$, such as:

$$T_O > F_O > I_O > T > F > I > I_U > F_U > T_U. \qquad (163)$$

# 70. Neutrosophic Offgraph

Let $V_j$, with $j \in \{1, 2, \dots, n\}$, and $n$ an integer, $n \geq 1$, be a set of vertices, and $E_{kl}$, with $k, l \in \{1, 2, \dots, n\}$ a set of edges that connect the vertex $V_k$ with the vertex $V_l$.

Each vertex $V_j$ has a neutrosophic membership degree of the form $V_j(T_j, I_j, F_j)$, with $T_j, I_j, F_j \subseteq [0, 1]$, and each edge $E_{kl}$ represents a neutrosophic relationship degree of the form $E_{kl}(T_{kl}, I_{kl}, F_{kl})$, with $T_{kl}, I_{kl}, F_{kl} \subseteq [0, 1]$.

Such graph is a **neutrosophic graph**.

Now, if there exists at least a vertex $V_{j_o}(T_{j_o}, I_{j_o}, F_{j_o})$ or at least an edge $E_{k_0 l_o}(T_{k_0 l_o}, I_{k_0 l_o}, F_{k_0 l_o})$, such that at least two of the neutrosophic components $T_{j_o}, I_{j_o}, F_{j_o}$, $T_{k_0 l_o}, I_{k_0 l_o}, F_{k_0 l_o}$ are partially or totally off the interval $[0, 1]$, one above and the other one below, then the graph

$$G_O = \{V_j, E_{kl}, \text{with } j, k, l \in \{1, 2, \dots, n\}, n \geq 1\} \quad (164)$$

is a **Neutrosophic Offgraph**.

# Example of Neutrosophic Offgraph

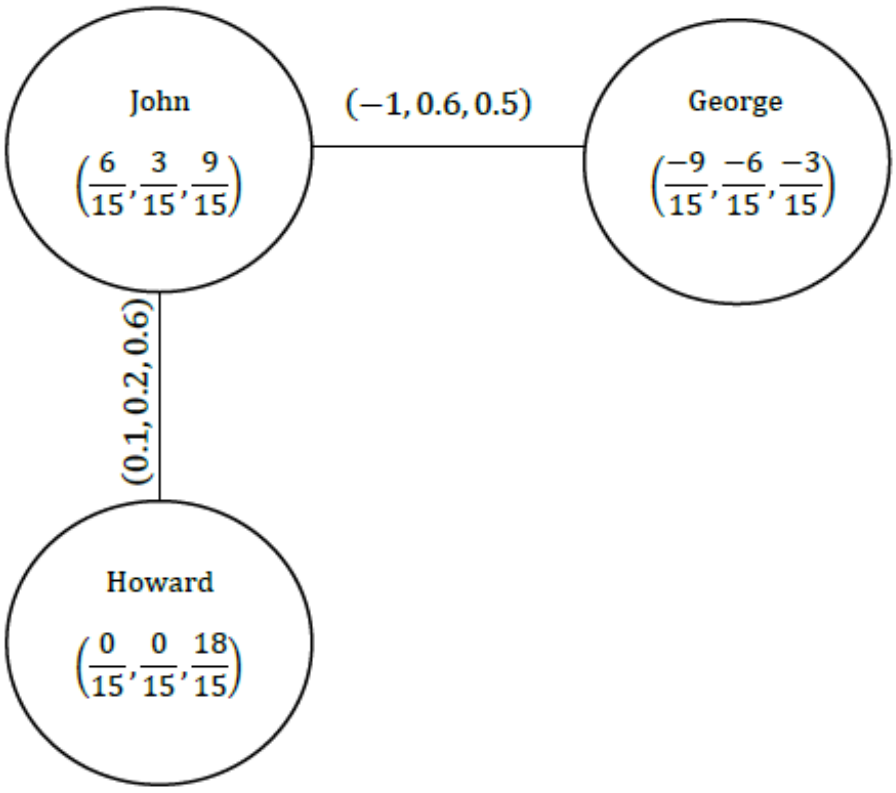

*Fig. 5*

since $\frac{18}{15} = 1.2 > 1$, also $\frac{-9}{15} < 0, \frac{-6}{15} < 0, \frac{-3}{15} < 0, -1 < 0$.

We reconsidered the previous example of enrollment of the students John, George, and Howard to the University Alpha as vertices, and we added some relationships between them.

# 71. Neutrosophic Bipolar/ Tripolar/ Multipolar Graph

We introduce for the first time the notions below.

**1) Neutrosophic Bipolar Graph**

Which is a graph that has the vertexes $V_j$ of the form (<$T^{+}{}_{j}$ , $T^{-}{}_{j}$ >, < $F^{+}{}_{j}$ , $F^{-}{}_{j}$ >), meaning their neutrosophic positive degree is <$T^{+}{}_{j}, I^{+}{}_{j}, F^{+}{}_{j}$> and their neutrosophic negative membership degree is $< T^{-}{}_{j}$ , $I^{-}{}_{j}$ , $F^{-}{}_{j} >$ with respect to the graph;

Edges $E_{jk}$ of the form (<$T^{+}{}_{jk}, T^{-}{}_{jk}$>, <$I^{+}{}_{jk}, I^{-}{}_{jk}$>, <$F^{+}{}_{jk}$, $F^{-}{}_{jk}$ >), meaning their neutrosophic positive relationship degree is $< T^{+}{}_{jk}$, <$I^{+}{}_{jk}$, $F^{+}{}_{jk} >$ between the vertexes $V_j$ and $V_k$ and their neutrosophic negative relationship is <$T^{-}{}_{jk}$, $I^{-}{}_{jk}$, $F^{-}{}_{jk}$>); or both.

2) If at least one of $T^{+}_{j_0}, I^{+}_{j_0}, F^{+}_{j_0}, T^{+}_{j_0k_o}, I^{+}_{j_0k_o}, F^{+}_{j_0k_o}$ for some given $j_0$∈ {1, 2, …, m} and $k_0$∈ {1, 2, …, p} is > 1, one has a
**Neutrosophic Bipolar Overgraph.**

3) Similarly, if at least one of $T^{-}_{j_1}, I^{-}_{j_1}, F^{-}_{j_1}, T^{-}_{j_1k_1}, I^{-}_{j_1k_1}, F^{-}_{j_1k_1}$, for some given $j_1$∈ {1, 2, …, m} and $k_1$∈ {1, 2, …, p}, is < - 1, one has a
**Neutrosophic Bipolar Undergraph**.

4) A neutrosophic bipolar graph which is both overgraph and undergraph is called a
**Neutrosophic Bipolar Offgraph**.

Example:

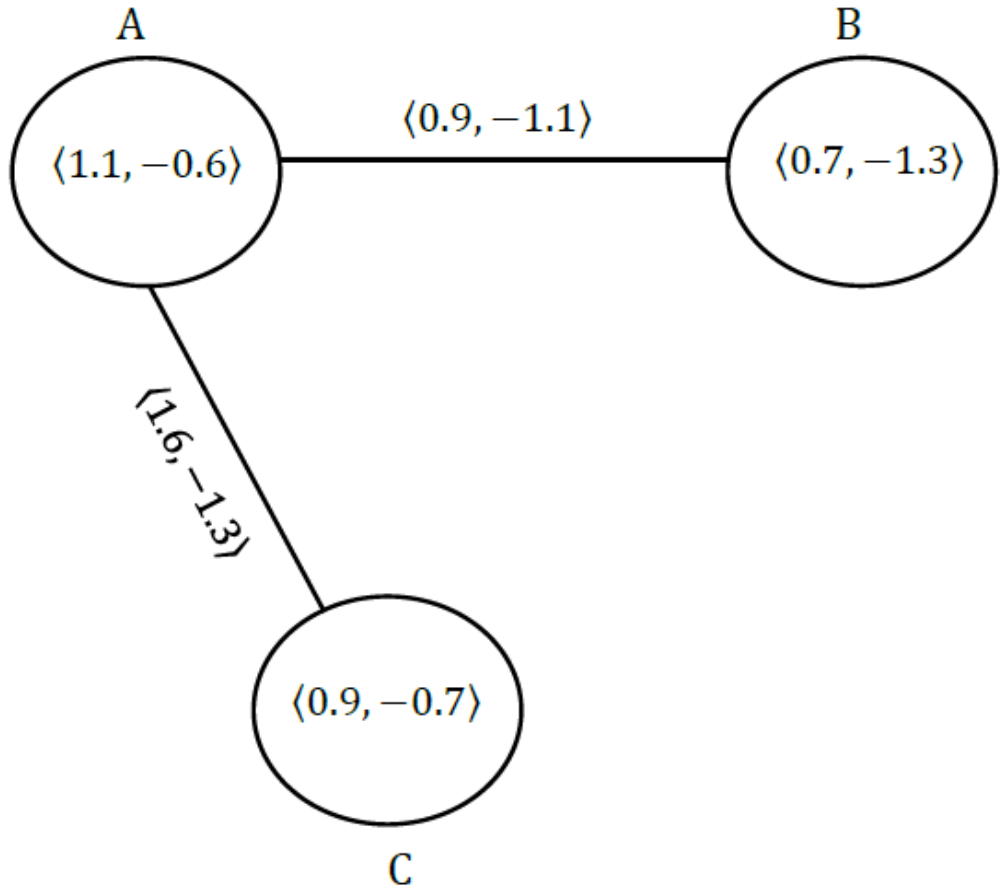

*Fig. 6*

5) **Neutrosophic Tripolar Graph** is a graph that has the vertexes $V_j$ of the form: (<$T^+{}_j$, $T^0{}_j$, $T^-{}_j$>, < $I^+{}_j$ , $I^0{}_j$, $I^-{}_j$> , < $F^+{}_j$ , $F^0{}_j$ , $F^-{}_j$ >) where < $T^+{}_j, I^+{}_j, F^+{}_j$ > is their neutrosophic positive membership degree, < $T^0{}_j$ , $I^0{}_j$ , $F^0{}_j$ > is their neutrosophic neutral membership degree, while < $T^-{}_j, I^-{}_j, F^-{}_j$ > is their negative membership degree, where for all j ∈{1, 2, …, m} one has:

$T^+{}_j\ I^+{}_j, F^+{}_j \subseteq [0, 1]$;

$T^-{}_j\ I^-{}_j, F^-{}_j \subseteq [-1, 0]$;

$T^0{}_j\ I^0{}_j, F^0{}_j \subseteq [-1, 1]$. (165)

One considers that the positive neutrosophic components are provides by a friendly source (which is biased towards positiveness), the negative neutrosophic components are providedby an enemy sources (which is biased towards negativeness), while the neutral neutrosophic components are provided by a neutral source (which is considered unbiased).

Similarly, the edges $E_{jk}$ have the form

$$(<T^{+}{}_{jk}, T^{0}{}_{jk}, T^{-}{}_{jk}>, <I^{+}{}_{jk}, I^{0}{}_{j}, I^{-}{}_{jk}>, <F^{+}{}_{jk}, F^{0}{}_{jk}, F^{-}{}_{jk}>) \quad (166)$$

representing their neutrosophic degrees of relationship between vertexes $V_j$ and $V_k$ :

where $< T^{+}{}_{jk}, I^{+}{}_{jk}, F^{+}{}_{jk} >$ is their neutrosophic positive relationship degree, $< T^{0}{}_{jk}, I^{0}{}_{jk}, F^{0}{}_{jk} >$ is their neutrosophic neutral relationship degree, while $< T^{-}{}_{jk}, I^{-}{}_{jk}, F^{-}{}_{jk} >$ is their negative relationship degree, where for all j ∈ {1, 2, …, m} and k ∈ {1, 2, …, p} one has:

$$T^{+}{}_{jk}\, I^{+}{}_{jk}, F^{+}{}_{jk} \subseteq [0, 1]; \quad (167)$$

$$T^{-}{}_{jk}\, I^{-}{}_{jk}, F^{-}{}_{jk} \subseteq [-1, 0]; \quad (168)$$

$$T^{0}{}_{jk}\, I^{0}{}_{jk}, F^{0}{}_{jk} \subseteq [-1, 1]. \quad (169)$$

6) The **Neutrosophic Tripolar Overgraph** has at least one positive neutrosophic component > 1.
7) The **Neutrosophic Tripolar Undergraph** has at least one neutrosophic component < -1.
8) The **Neutrosophic Tripolar Offgraph** has both: a positive neutrosophic component > 1, and a negative neutrosophic component < -1.
9) The **Neutrosophic Multipolar Graph** is a grapg that has the vertexes $V_j$ whose neutrosophic membership degress have the forms of

neutrosophic multipolar sets, or the edges $E_{jk}$ whose relationship degrees have the forms of neutrosophic multipolar sets.

10) Similarly, the **Neutrosophic Multipolar Overgraph** has at least a vertex or an edge characterized by a neutrosophic multipolar overset.
11) The **Neutrosophic Multipolar Undergraph** has at least a vertex or an edge characterized by a neutrosophic multipolar underset.
12) The **Neutrosophic Multipolar Offgraph** includes both, the neutrosophic multipolar overgraph and the neutrosophic multipolar undergraph.

# 72. Neutrosophic Bipolar *(t, i, f)*- Matrix

We introduce for the first time the notions of **Neutrosophic Bipolar *(t, i, f)*-Matrix**, which is a matrix $M$ that has at least one element $x \in \mathcal{U}$ of Neutrosophic Bipolar form, i.e.

$$x(< T_x^+, T_x^- >, < I_x^+, I_x^- >, < F_x^+, F_x^- >), \qquad (170)$$

where $T_x^+, I_x^+, F_x^+$ are the positive degrees of membership, indeterminate-membership, and nonmembership with respect to the matrix respectively included in [0,1], and $T_x^-, I_x^-, F_x^-$ are the negative degrees of the membership, indeterminate-membership, and nonmembership respectively included in [-1,0].

In general, we consider a neutrosophic bipolar set $A \subset \mathcal{U}$ and a matrix, whose elements are neutrosophic bipolar numbers from $A$. Then the matrix $M$ is a neutrosophic bipolar matrix.

## Example of Neutrosophic Bipolar *(t, i, f)*-Matrix

$$M_1 = \begin{bmatrix} 4(< 0.9, -0.1 >, < 0.1, -0.2 >, < 0.0, -0.3 >) & 5(< 0.2, -0.2 >, < 0.5, -0.3 >, < 0.6, -0.5 >) \\ 7(< 0.1, -0.6 > < 0.5 \ - 0.5 >, < 0.2, -0.2 >) & 8(< 0.1, -0.1 >, < 0.4, -0.3 >, < 0.3, -0.2 >) \end{bmatrix}$$

A **Neutrosophic Bipolar $(\boldsymbol{t}, \boldsymbol{i}, \boldsymbol{f})$ -Overmatrix** is a neutrosophic bipolar matrix that has at least one element $x_1 \in U$ with a positive degree among $T_{x_1}^+$, $I_{x_1}^+$, $F_{x_1}^+$, that is partially or totally above 1. An example of such element: $x_1$ (<1.5, -0.1>, <0.0, -0.4>, <0.1, -0.2>), where $T_{x_1}^+ = 1.5 > 1$.

A **Neutrosophic Bipolar $(\boldsymbol{t}, \boldsymbol{i}, \boldsymbol{f})$ -Undermatrix** is a neutrosophic bipolar matrix that has at least one element $x_2 \in U$ with a negative degree among $T_{x_2}^{+}$, $I_{x_2}^{+}$, $F_{x_2}^{+}$, that is partially or totally below -1.

Example of such element: $x_2$ (<0.2, -0.4>, <0.0, -0.3>, <[0.2, 0.4], [-1.3, -0.5]>), where $F_{x_2}^{+} = [-1.3, -0.5]$ is partially bellow -1.

A **Neutrosophic Bipolar** $(\boldsymbol{t}, \boldsymbol{i}, \boldsymbol{f}) -$ **Offmatrix** is a matrix that is both a neutrosophic bipolar overmatrix and a neutrosophic bipolar undermatrix.

## Examples of Neutrosophic Bipolar *(t, i, f)*-Offmatrix

$M_2 = \begin{bmatrix} 5_{(\langle 1.7,-0.2\rangle,\langle 0.1,-0.3\rangle,\langle 0.2,-0.1\rangle)} \\ 9_{(\langle 0.4,-0.1\rangle,\langle 0.0,-0.1\rangle,\langle 0.5,-1.6\rangle)} \end{bmatrix}$ of size $2 \times 1$.

Also,

$M_3 = \left[47_{(\langle 0.2,-1.2\rangle,\langle 1.3,-0.1\rangle,\langle 0.0,-0.5\rangle)}\right]$ of size $1 \times 1$, since $I_{47}^{+} = 1.3 > 1$ and $T_{47}^{-} = -1.2 < -1$.

# 73. Neutrosophic Tripolar *(t, i, f)*-Matrix

Neutrosophic Tripolar *(t, i, f)*-Matrix is a matrix that contains at least one element $x \in \mathcal{U}$ of neutrosophic tripolar form, i.e.

$$x(\langle T_x^+, T_x^O, T_x^- \rangle, \langle I_x^+, I_x^O, I_x^- \rangle, \langle F_x^+, F_x^O, F_x^- \rangle), \qquad (171)$$

where

$T_x^+, I_x^+, F_x^+ \subseteq [0, 1]$ are positive degrees of membership, indeterminate-membership, and nonmembership, with respect to the matrix [provided by a friendly source];

$T_x^-, I_x^-, F_x^- \subseteq [-1, 0]$ are negative degrees of membership, indeterminate-membership, and nonmembership, with respect to the matrix [provided by an enemy source];

$T_x^O, I_x^O, F_x^O \subseteq [-1, 1]$ are neutral degrees of membership, indeterminate-membership, and nonmembership, with respect to the matrix [provided by a neutral source].

## Example of a Neutrosophic Tripolar Element

$x(\langle 0.6, 0.4, -0.1 \rangle, \langle 0.2, 0.1, -0.3 \rangle, \langle 0.4, 0.6, 0.0 \rangle)$.

# 74. Neutrosophic Tripolar *(t, i, f)*-Overmatrix

A **Neutrosophic Tripolar *(t, i, f)*-Overmatrix** is a matrix that contains at least a neutrosophic tripolar element $x \in \mathcal{U}$ such that at least one of its positive or neutral neutrosophic components $T_x^+, I_x^+, F_x^+, T_x^0, I_x^0, F_x^0$ is partially or totally above 1. This is called a **neutrosophic tripolar overelement**.

## Example of such element

$x(\langle 0.6, 0.1, -0.2\rangle, \langle 0.2, 0.7, -0.6\rangle, \langle 0.4, 1.6, -0.6\rangle)$,
since $F_x^0 = 1.6 > 1$.

# 75. Neutrosophic Tripolar *(t, i, f)*-Undermatrix

A **Neutrosophic Tripolar *(t, i, f)*-Undermatrix** is a matrix that contains at least a neutrosophic tripolar element $x \in \mathcal{U}$ such that at least one of its negative or neutral components $T_x^-, I_x^-, F_x^-, T_x^0, I_x^0, F_x^0$ is partially or totally below $-1$. This is called a **neutrosophic tripolar underelement**.

## Example of such element

$$x(\langle 0.5, 0.5, -1.7\rangle, \langle 0.1, -0.2, 0.0\rangle, \langle 0.1, (-1.1, -1), -0.3\rangle)$$

since $T_x^- = -1.7 < -1$, and also $F_x^0(-1, 1, -1)$ is totally below $-1$.

# 76. Neutrosophic Tripolar *(t, i, f)*-Offmatrix

A **Neutrosophic Tripolar *(t, i, f)*-Offmatrix** is a matrix that contains: either a neutrosophic tripolar overelement and a neutrosophic tripolar underelement, or a neutrosophic tripolar offelement.

A **Neutrosophic Tripolar *(t, i, f)*-offelement** is an element $x \in \mathcal{U}$ such that it has among its 9 neutrosophic subcomponents: at least one which is partially or totally above 1, and another one which is partially or totally below $-1$.

## Example of a Neutrosophic Tripolar *(t, i, f)*-Offelement

$x(\langle[1.0,1.2], 0.0, -0.7\rangle, \langle 0.1, -0.2, -0.3\rangle, \langle 0.2, 0.4, -1.3\rangle)$, since $T_x^+ = [1.0, 1.2]$ is partially above 1, and $F_x^- = -1.3 < -1$.

# 77. *(t, i, f)*-Neutrosophic Over-/ Under-/ Off-Matrix

In the **classical matrix theory** $M = \left(a_{j_k}\right)_{j_k,}$ where $j \in \{1, 2, \ldots, m\}, k \in \{1, 2, \ldots, n\}$, with $j, k \geq 1$, and all $a_{j_k} \in \mathbb{R}$, each element belongs to the matrix 100%. For example:

$$A = \begin{bmatrix} 2 & 5 \\ -1 & 0 \end{bmatrix},$$

which can be translated in a neutrosophic way as:

$$A_N = \begin{bmatrix} 2_{(1,0,0)} & 5_{(1,0,0)} \\ -1_{(1,0,0)} & 0_{(1,0,0)} \end{bmatrix},$$

meaning that each element belongs to the matrix 100%, its indeterminate-membership 0%, and its nonmembership degree is 0%.

But in our reality, there are elements that only partially belong to a set, or to a structure, or to an entity, generally speaking.

We introduce for the first time the *(t, i, f)*-neutrosophic matrix, which is a matrix that has some element that only partially belongs to the matrix:

$$M_N = \left(a_{j_k(t_{jk}, i_{jk}, f_{jk})}\right) jk, \qquad (172)$$

which means that each element $a_{j_k}$ belongs in a $(t_{jk}, i_{jk}, f_{jk})$ neutrosophic way to the matrix, i.e. $t_{jk}$ is its membership degree, $i_{jk}$ is its indeterminate-membership degree, and $f_{jk}$ is its nonmembership degree.

## Example

$$B_N = \begin{bmatrix} 4_{(-0.1,0.2,0.5} & -2_{(0.8,0.1,0.1)} \\ 3_{(0.6,0.0,0.7)} & 1_{(0.7,0.1,0.0)} \end{bmatrix}.$$

We call it "*(t, i, f)*-neutrosophic matrix", in order to distinguish it from the previous "neutrosophic matrix" defined on numbers of the form $a + bI$, where $I$ = indeterminacy, and $I^2 = I$, while *a, b* are real or complex numbers.

For example:

$$C = \begin{bmatrix} 2 & I & 3 \\ -4I & 0 & 1 \end{bmatrix}$$

is just a neutrosophic matrix.

*

We now introduce for the first time the following three new notions:

1. ***(t, i, f)*-Neutrosophic Overmatrix**, which is a *(t, i, f)*-neutrosophic matrix such that at least one of its elements has at least one neutrosophic component that is partially or totally above 1.

For example:

$$D_N = \begin{bmatrix} 21_{(0.1,0.3,[0.9,1.1])} & 33_{(0.6,(0.7,0.8),0.9)} \\ 7_{(1,0,0)} & -5_{(0,0,1)} \end{bmatrix},$$

since the interval $[0.9, 1.1]$ is partially above 1.

2. ***(t, i, f)*-Neutrosophic Undermatrix**, which is a *(t, i, f)*-neutrosophic matrix such that at least one of its elements has at least one neutrosophic component that is partially or totally below 0.

For example:

$$E_N = [0_{(1,0,1)} \quad -2_{(0.2,[0.1,0.3],\{-0.3,0.0\})}],$$

because $\{-0.3, 0.0\}$ is partially below 0 since $-0.3 < 0$.

3. ***(t, i, f)*-Neutrosophic Offmatrix**, which is a *(t, i, f)*-neutrosophic matrix such that at least one of its elements has at least one component that is partially or totally above 1, and at least one component of this element that is partially or totally below 0.

For example:

$G_N = [25_{(-0.1,0.2,1.3)} \quad 23_{(0,1,0)} \quad 51_{(0.2,(-0.1,0.1),0.8)}]$.

# 78. Complex Neutrosophic Set

Complex Neutrosophic Set $S_N$ {presented first time by Ali and Smarandache in 2015} on a universe of discourse U, is defined as:

$$S_N = \{(x, < t_1(x)e^{j \cdot t_2(x)}, i_1{}^{j \cdot i_2(x)} f_1(x)e^{j.f2(x)} >), x \in U\} \quad (173)$$

where $t_1(x)$ is the amplitude membership degree,

$t_2(x)$ is the phase membersip degree,

$i_1(x)$ is the amplitude indeterminate-membership degree,

$i_2(x)$ is the phase indeterminate-membership degree,

$f_1(x)$ is the amplitude nonmembership degree,

$f_2(x)$ is the phase nonmembership degree of the element x with respect to the neutrosophic set $S_N$, where $t_1(x)$, $i_1(x), f_1(x)$ are standard or non-standard subsets of the non-standard unit-interval $]^-0, 1^+[$, while $t_2(x), i_2(x), f_2(x)$ are subsets of the set of real numbers $\mathbb{R}$. This is the most general definition of the complex neutrosophic set. The non-standard subsets are used only to make distinction between "absolute" and "relative" truth, indeterminacy or falsehood in philosophy. A truth (or indeterminacy, or falsehood) is absolute if it occurs in all possible worlds (Leinitz), and relative if it occurs in at least one world. Since in science and technology we do not need "absolute" or "relative", we'll be working only with standard real subsets, and with the standard real interval [0,1]. Particular cases can be studied, like:

**Complex Neutrosophic Overset** that is a complex neutrosophic set that has for at least one element x ∈ U, such that at least one of its neutrosophic subcomponents $t_1$(x), $t_2(x)$, $i_1(x)$, $i_2(x)$, $f_1(x)$, $f_2(x)$ is partially or totally > 1.

For example: let U be a universe of discourse. Then $A =$ {$x_1$ ($1.2e^{j.\pi}$, 0.7 $e^{\frac{j\pi}{2}}$, $0.1e^{j2\pi}$)}, $x_2$ ($0.6e^{j(2.6)}$, [0.9, 1.1]$e^{j.5}$, $0.5e^{j.3}$); $x_1$ , $x_2$ ∈ U} is a complex neutrosophic overset, since $t_1^{x_1}$ = 1.2 > 1, also $i_2^{x_2}$ = [0.9, 1.1] is partially above 1.

**Complex Neutrosophic Underset** is a complex neutrosophic set that has at least one element x ∈ U, such that at least one of its neutrosophic subcomponents $t_1$(x), $t_2(x)$, $i_1(x)$, $i_2(x)$, $f_1(x)$, $f_2(x)$ is partially or totally < 1.

For example:

B= {$x_1$ ($0.7e^{j.3}$, [0.6, 0.7)$e^{j\cdot[4,5]}$, (-0.8, 0)$e^{j.3}$), $x_1$ ∈ U} is a complex Neutrosophic underset since ${f_1}^{x_1}$ = (-0.8, 0) is totally below 0 (zero).

**Complex Neutrosophic Offset** is a complex neutrosophic set that has at least one neutrosophic subcomponent among $t_1(x)$, $t_2(x)$, $i_1(x)$, $i_2(x)$, $f_1(x)$, $f_2(x)$ partially or totally > 1 for some element x ∈ U and at least one neutrosophic subcomponent among $t_1$(y), $t_2(y)$, $i_1(y)$, $i_2(y)$, $f_1(y)$, $f_2(y)$ partially or totally < 0 for some element y ∈ U.

For examples: C = {$x_1$ (0.2 $e^{j\cdot(4.2)}$, 0.1$e^{j\cdot(4.2)}$, [0.8, 15]$\cdot$ $e^{j\cdot[0.8,0.9]}$), $x_2$(-0.6$e^{j\cdot(0.9)}$, 0.2$e^{j\cdot(4)}$, 1$\cdot e^{j\cdot(5)}$ ), $x_1$, $x_2$∈ U}, because of ${f_1}^{x_1}$ = [0.8, 1.5] is partially above 1, and ${t_1}^{x_2}$= -0.6 < 0.

D={$x_3$(-0.7$e^{j\cdot(7)}$, 0.6$e^{j\cdot(2)}$, 1.3$e^{j\cdot(9)}$), $x_3$∈ U} since ${t_1}^{x_3}$ = -0.7 < 0 and ${f_1}^{x_3}$ = 1.3 > 1.

# 79. General Neutrosophic Overtopology

Let's consider a universe of discourse U, and a non-empty neutrosophic offset $M_O \subset$ U.

A General Neutrosophic Overtopology on $M_O$ is a family $\eta_O$ that satisfies the following axioms:

a) $0(0, \Omega_I, \Omega_F)$ and $\Omega_T(\Omega_T, 0, 0) \in \eta_O$, where $\Omega_T$ is the overtruth (highest truth-value, which may be > 1), $\Omega_I$ is the overindeterminacy (highest indeterminate-value, which may be > 1), and $\Omega_F$ is the overfalsehood (highest falsehood-value, which may be > 1); at least one of $\Omega_T, \Omega_I, \Omega_F$ has to be >1 in order to deal with overtopology.

b) If A, B $\in \eta_O$, then A $\cap B \in \eta_O$.

c) If the family $\{A_k$, k $\in$ K$\} \subset \eta_O$, then $\cup_{k\in K} A_k \in \eta_O$.

# 80. General Neutrosophic Undertopology

**General Neutrosophic Undertopology** on the neutrosophic underset $M_U$, included in U, is defined in a similar way, as a family $\eta_U$ except the first axiom which is replaced by:

a) $\Psi_T(\Psi_T, 1, 1)$ and $1(1, \Psi_I, \Psi_F) \in \eta$, (174)

where $\Psi_T$ is the undertruth (lowest truth-value, which may be < 0), and $\Psi_I$ is the underindetereminacy (lowest indeterminacy-value which may be < 0), and $\Psi_F$ is the underfalsehood (lowest falsehood-value which may be <0 ); at least one of $\Psi_T, \Psi_I, \Psi_F$ has to be < 0 in order to deal with undertopology.

# 81. General Neutrosophic Offtopology

**General Neutrosophic Offtopology** on the neutrosophic offset $M_{off} \subset$ U, is defined similarly as a family $\eta_{off}$ of neutrosophic (off)sets in $M_{off}$, again except the first axiom which is replaced by:

a) $\Psi_T(\Psi_T, \Omega_I, \Omega_F)$ and $\Omega_T(\Omega_T, \Psi_I, \Psi_F) \in \eta_{off}$. (175)

## *Author's Presentations at Seminars and National and International Conferences on Neutrosophic Over- / Under- / Off -Set, -Logic, -Probability, and -Statistics*

The author has presented the

- *neutrosophic overset, neutrosophic underset, neutrosophic offset;*
- *neutrosophic overlogic, neutrosophic underlogic, neutrosophic offlogic;*
- *neutrosophic overmeasure, neutrosophic undermeasure, neutrosophic offmeasure;*
- *neutrosophic overprobability, neutrosophic underprobability, neutrosophic offprobability;*
- *neutrosophic overstatistics, neutrosophic understatistics, neutrosophic offstatistics;* as follows:

*Statistics / Neutrosophic Precalculus and Calculus / Symbolic Neutrosophic Theory / Open Challenges of Neutrosophic Set*, Vietnam national University, Vietnam Institute for Advanced Study in Mathematics, Hanoi, Vietnam, lecture series, 14th May – 26th May 2016.

17. *Foundations of Neutrosophic Logic and Set and their Applications to Information Fusion,* Hanoi University, 18th May 2016.
18. *Neutrosophic Theory and Applications,* Le Quy Don Technical University, Faculty of Information Technology, Hanoi, Vietnam, 17th May 2016.
19. *Types of Neutrosophic Graphs and Neutrosophic Algebraic Structures together with their Applications in Technology*, Universitatea Transilvania din Brasov, Facultatea de Design de Produs si Mediu, Brasov, Romania, 6 June 2015.
20. *Foundations of Neutrosophic Logic and Set and their Applications to Information Fusion*, tutorial, by Florentin Smarandache, 17th International Conference on Information Fusion, Salamanca, Spain, 7th July 2014.
21. *Foundations of Neutrosophic Set and Logic and Their Applications to Information Fusion*, by F. Smarandache, Osaka University, Inuiguchi Laboratory, Department of Engineering Science, Osaka, Japan, 10 January 2014.
22. *Foundations of Neutrosophic set and Logic and Their Applications to Information Fusion*, by F. Smarandache, Okayama University of Science, Kroumov Laboratory, Department of Intelligence Engineering, Okayama, Japan, 17 December 2013.
23. *Fundations of Neutrosophic Logic and Set and their Applications to Information Fusion*, by Florentin Smarandache, Institute of Extenics Research and Innovative Methods, Guangdong University of Technology, Guangzhou, China, July 2nd, 2012.
24. *Neutrosophic Logic and Set Applied to Robotics*, seminar to the Ph D students of the Institute of Mechanical Solids of the Romanian Academy, Bucharest, December 14, 2011.
25. *Foundations and Applications of Information Fusion to Robotics*, seminar to the Ph D students of the Institute of

*Neutrosophic Over-/Under-/Off-Set and -Logic* were defined for the first time by Smarandache in 1995 and published in 2007. They are totally different from other sets/logics/probabilities.

He extended the neutrosophic set respectively to *Neutrosophic Overset* {when some neutrosophic component is > 1}, *Neutrosophic Underset* {when some neutrosophic component is < 0}, and to *Neutrosophic Offset* {when some neutrosophic components are off the interval [0, 1], i.e. some neutrosophic component > 1 and other neutrosophic component < 0}.

This is no surprise with respect to the classical fuzzy set/logic, intuitionistic fuzzy set/logic, or classical/imprecise probability, where the values are not allowed outside the interval [0, 1], since our real-world has numerous examples and applications of over-/under-/off-neutrosophic components.

*Example of Neutrosophic Offset.*

In a given company a full-time employer works 40 hours per week. Let's consider the last week period.

Helen worked part-time, only 30 hours, and the other 10 hours she was absent without payment; hence, her membership degree was 30/40 = 0.75 < 1.

John worked full-time, 40 hours, so he had the membership degree 40/40 = 1, with respect to this company.

But George worked overtime 5 hours, so his membership degree was (40+5)/40 = 45/40 = 1.125 > 1. Thus, we need to make distinction between employees who work overtime, and those who work full-time or part-time. That's why we need to associate a degree of membership strictly greater than 1 to the overtime workers.

Now, another employee, Jane, was absent without pay for the whole week, so her degree of membership was 0/40 = 0.

Yet, Richard, who was also hired as a full-time, not only didn't come to work last week at all (0 worked hours), but he produced, by accidentally starting a devastating fire, much damage to the company, which was estimated at a value half of his salary (i.e. as he would have gotten for working 20 hours that week). Therefore, his membership degree has to be less that Jane's (since Jane produced no damage). Whence, Richard's degree of membership, with respect to this company, was - 20/40 = - 0.50 < 0.

Consequently, we need to make distinction between employees who produce damage, and those who produce profit, or produce neither damage no profit to the company.

Therefore, the membership degrees > 1 and < 0 are real in our world, so we have to take them into consideration.

Then, similarly, the Neutrosophic Logic/Measure/Probability/Statistics etc. were extended to respectively *Neutrosophic Over-/Under-/Off-Logic, -Measure, -Probability, -Statistics* etc. [Smarandache, 2007].